\documentclass[journal]{IEEEtran}
\usepackage{color}
\usepackage{times}
\usepackage{epsfig}
\usepackage{graphicx}
\usepackage{amsmath}
\usepackage{amssymb}
\usepackage{bm}
\usepackage{subfigure}
\usepackage[pagebackref=true,breaklinks=true,letterpaper=true,colorlinks,bookmarks=false]{hyperref}
\usepackage{amsmath,amssymb}
\usepackage[ruled,vlined]{algorithm2e}
\usepackage{multirow}
\usepackage{multicol}
\usepackage{array}

\newcommand{\etal}{\textit{et al}. }

\begin{document}

\title{Improving Person Re-identification with Iterative Impression Aggregation}

\author{Dengpan Fu*, Bo Xin,~\IEEEmembership{Member,~IEEE}, Jingdong Wang,~\IEEEmembership{Senior Member,~IEEE}, Dongdong Chen, Jianmin Bao, Gang Hua,~\IEEEmembership{Fellow,~IEEE} and Houqiang Li,~\IEEEmembership{Senior Member,~IEEE}
\thanks{Dengpan Fu and Houqiang Li are with the Department of Electronic Engineering and Information Science, University of Science and Technology of China. (e-mail: fdpan@mail.ustc.edu.cn,lihq@ustc.edu.cn). This work is done when Dengpan Fu is an intern at Microsoft Research Asia.}
\thanks{Bo Xin, Jianmin Bao and Jingdong Wang are with the Microsoft Research, Beijing, China. (e-mail: jimxinbo@gmail.com, jianbao@microsoft.com, jingdw@microsoft.com.)}
\thanks{Dongdong Chen is with Microsoft Cloud AI, Redmond, US. (e-mail: cddlyf@gmail.com.)}
\thanks{Gang Hua is with Wormpex AI Research, Bellevue, WA, USA. (email: ganghua@gmail.com.)}
}

\markboth{Journal of \LaTeX\ Class Files,~Vol.~14, No.~8, August~2020}%
{Shell \MakeLowercase{\textit{et al.}}: Bare Demo of IEEEtran.cls for IEEE Journals}

\maketitle

\begin{abstract}
Our impression about one person often updates after we see more aspects of him/her and this process keeps iterating given more meetings. We formulate such an intuition into the problem of person re-identification (re-ID), where the representation of a query (probe) image is iteratively updated with new information from the candidates in the gallery. Specifically, we propose a simple attentional aggregation formulation to instantiate this idea and showcase that such a pipeline achieves competitive performance on standard benchmarks including CUHK03, Market-1501 and DukeMTMC. Not only does such a simple method improve the performance of the baseline models, it also achieves comparable performance with latest advanced re-ranking methods. Another advantage of this proposal is its flexibility to incorporate different representations and similarity metrics. By utilizing stronger representations and metrics, we further demonstrate state-of-the-art person re-ID performance, which also validates the general applicability of the proposed method.
\end{abstract}


\IEEEpeerreviewmaketitle

\section{Introduction}
\label{sec:intro}

In video surveillance, when presented with a query person of interest, person re-ID aims to arbitrate whether the person has been observed by another camera in a different place or time. Due to the increasing number of camera networks and a surging demand of public safety, person re-ID has become an increasingly demanding task in computer vision and drawn a lot of attention from both academia and industry.

\begin{figure}[!ht]
    \centering
    \includegraphics[width=2.8in]{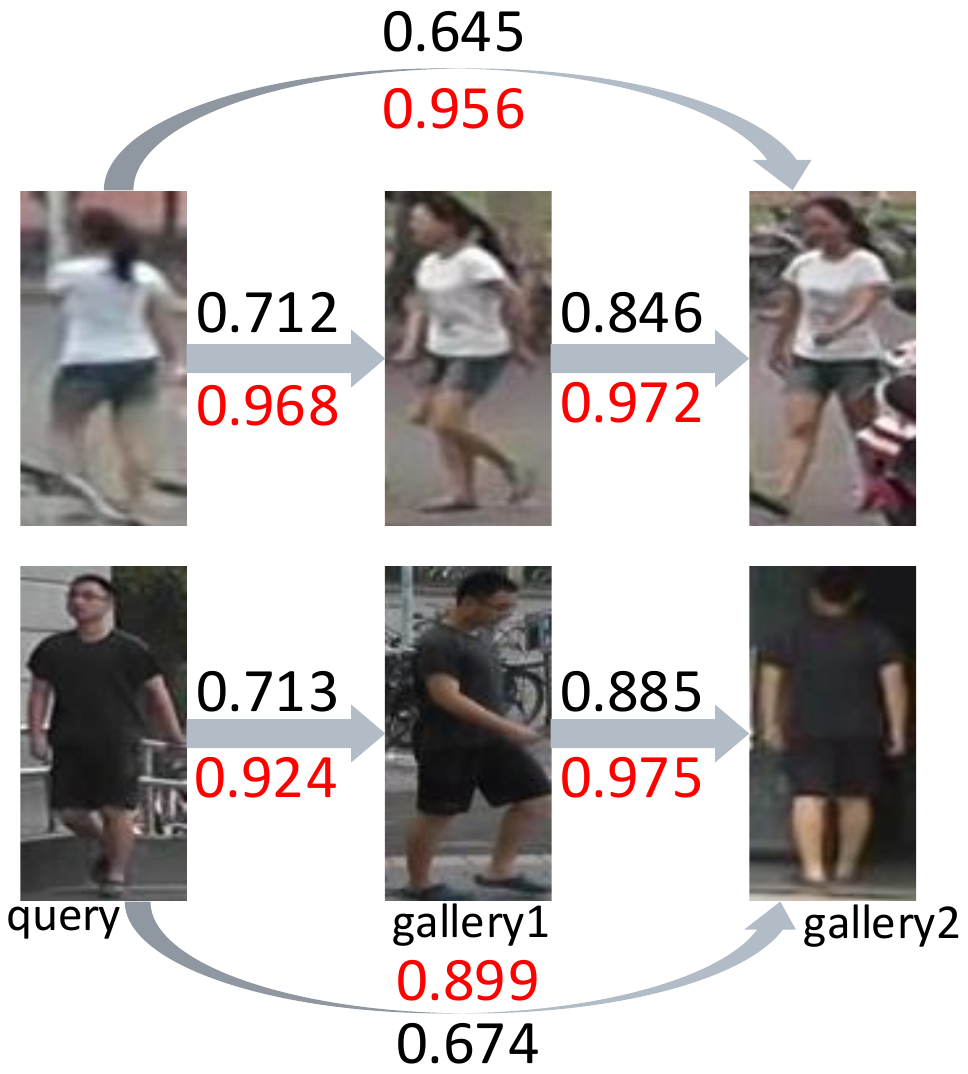}
    \caption{Illustration of aggregating gallery information. The numbers show the value of a similarity measurement (the higher the more similar), numbers in red represent updated similarities after impression aggregation. In both cases, the original similarity between the query image and gallery2 is relatively low, but since the query image is close to gallery1 and after identifying gallery1 and aggregating information from gallery1, the new representation/impression of the query becomes more similar to gallery2.}
    \label{fig:example}
\end{figure}

In this regard, contributions have been made mainly to building stronger person description/representation  \cite{wang2007shape,liu2017end,sun2018PCB,chen2018group,shen2018deep,suh2018part,zheng2019joint,eom2019learning} and better similarity metrics  \cite{koestinger2012large,weinberger2009distance,xiong2014person,liao2015person,martinel2015kernelized,paisitkriangkrai2015learning,oh2016deep,yang2017person} etc. Especially, recent deep learning based methods \cite{sun2018PCB,suh2018part,wang2018learning,zhang2020ordered} have achieved very promising performance on standard benchmarks.
However, due to the limited size of current datasets for person re-ID, it seems we are approaching a certain performance saturation after exhaustively exploring existing training data towards stronger representations and similarity metrics.

To this end, we propose to exploit more information from the gallery dataset at the retrieval phase to improve re-ID performance. Note that, since the gallery dataset often contains more images of the query person of interest, during the retrieval phase, images from the gallery of easy positive candidates could be explored to provide auxiliary information to the query image to better locate hard positive candidates. One example would be as follows: when searching for a person-of-interest, if the query image is of the person's frontal-view $F$ and the gallery has images of both the person's side-view $S$ and back-view $B$. Although matching $B$ directly to $F$ is difficult, we may be able to match $F$ to $S$ first and then leverage $S$ as a bridge to match $B$, so as to establish the match between $F$ and $B$. Fig. \ref{fig:example} illustrates two instances from real data, where updating the impression of the query with gallery information helps to find hard positive candidates more easily.

Recently, there is a surging attention to migrating re-ranking techniques of generic instance retrieval \cite{jegou2007contextual,qin2011hello,ma2015cross,bai2016sparse,garcia2017discriminant,zhong2017re} to the re-ID community due to their effectiveness in improving performance. Re-ranking methods indeed help to explore more information from the gallery dataset. Notwithstanding the demonstrated success of conventional re-ranking method by directly manipulating the retrieval ranks, we find alternatively that adaptively updating the query representation with relevant gallery information to be very effective in achieving better accuracy. This is partially inspired by the human visual system, which formulates an impression or description of the person-of-interest and keeps updating this impression when we see more aspects of this person \cite{mende2013neural}. 

Therefore, we propose a surrogate way to exploit gallery information, where we keep track of the representation of a query (probe) image and iteratively update it with new information from the candidates in the gallery.
At each iteration, the new representation, bearing the newly formed impression, could be used to compute similarities and retrieve instances of the same ID. 
Such a process can be viewed as one form of query expansion \cite{salton1990improving}, another classic idea from the retrieval literature, applied to person re-ID in an iterative manner.
We further note that, such an impression update could be applied to both the query image and all the images in the gallery, where we keep track of the impressions of all candidates as well.

We propose a simple iterative attentional aggregation model to formulate and computationally put such intuition into practice. Specifically, at each iteration, we first compute a similarity map between the query image and all the images from the gallery based on the current representations. Then we utilize the similarities to re-weight (similar candidates with higher weights and vice versa) and aggregate different gallery representations to update the query representation. Similar updates could also be optionally applied to all gallery images. Therefore, after each iteration, we can compute new similarities and retrieve instances based on the newly formed representations/impressions. Such an impression-update process keeps iterating until a saturating point, and the final representation will be used to facilitate retrieval. We name our proposed model  iterative impression aggregation (IIA) and refer to it as IIA in the rest of the paper.

Although the idea is simple and intuitive, our design of IIA inherently enjoys  flexibility to be combined with different representations and similarity metrics. In order to evaluate its effectiveness, extensive experiments have been conducted on standard person re-ID benchmarks including CUHK03, Market-1501 and DukeMTMC. We observe that a simple naive implementation with regular similarity metric could already largely improve the performance of competitive baseline models. It also achieves comparable performance with recent advanced re-ranking methods. With incorporation of stronger baseline representations and similarity metrics, we further demonstrate state-of-the-art performance.

In the remaining sections, we start by discussing related works in Section \ref{sec:related}. Then the detailed mathematical formulations and analysis of IIA will be elaborated in Section \ref{sec:method}. For better understanding, we also provide extensive ablation analysis in Section \ref{sec:expr}. Finally, we conclude this paper with discussions about future efforts in Section \ref{sec:concl}. Before proceeding, we summarize our key contributions as follows:

\begin{itemize}
    \item We propose an iterative impression aggregation model to improve person re-ID, which is an intuitive and novel way to explore gallery information for re-ID.
    \item We propose an effective attentional aggregation formulation, which is flexible to leverage different representations and similarity metrics for different purposes.
    \item We achieve state-of-the art performance on standard re-ID benchmarks, providing a simple practical post-processing module to strengthen the performance of existing re-ID methods.
\end{itemize}

\section{Related Work}
\label{sec:related}

Person re-ID remains to be one of the most active research area in computer vision with wide applications. We refer interested readers to \cite{zheng2016person} for a comprehensive review of the literature and here we mainly focus on discussions about recent research works that highly relate to our method.

\subsection{Representative baselines} Many works have been devoted to building stronger representations \cite{hermans2017defense,sun2018PCB,suh2018part,wang2018learning} and similarity metrics \cite{koestinger2012large,liao2015person} or in an end-to-end fashion to improve the performance. In \cite{sun2018PCB,suh2018part},  part-level features have been studied for better representation learning. In details, discriminative part-informed features with classification loss and part-aligned bi-linear representations are learned in \cite{sun2018PCB} and \cite{suh2018part} respectively. 
Alexander \etal \cite{hermans2017defense} revisited the importance of triplet loss and demonstrated that good representations can also be learnt with only triplet loss. In \cite{wang2018learning}, both part based classification loss and triplet loss are applied to learn better features and achieves new state-of-the-art performance. For metric learning, based on statistical inference, a distance metric is learnt with equivalence constraints in \cite{koestinger2012large}. And Liao \etal \cite{liao2015person} propose to learn discriminant metric by learning low dimensional subspace with cross-view quadratic discriminant analysis. Our proposed IIA is orthogonal with these works and aims to improve the performance from another aspect in better exploiting information in the gallery database. As we empirically observe, this is highly complementary to these baseline methods and our method often improves such baselines by a clear margin.

\subsection{Exploiting gallery information} To better model the relationship between the query and gallery, many works start to integrate the extra gallery information either in the training or inference stage. For the training stage integration, representative works include  \cite{chen2018group,shen2018deep,shen2018person}. Specifically, Chen \etal utilize conditional random field (CRF) for end-to-end group consistent similarities learning. And in  \cite{shen2018deep}, a different affinity metric between  gallery images is explored via random walk. To pass the message from gallery to query in a more structural way, a similarity-guided graph neural network is proposed in \cite{shen2018person}. By integrating gallery information, these methods can produce more compact and discriminative features.

Compared to training stage integration, gallery information exploitation in inference stage can also be viewed as a flexible post-processing step in many existing methods.  By exploring extra statistics from the testing set in the inference stage, they can gradually improve the performance  with negligible cost.  Existing methods can be roughly divided into two types: re-ranking and query expansion. Originally studied in generic instance retrieval  \cite{ye2016person,leng2015person}, re-ranking has drawn increasing attention in the re-ID community \cite{garcia2015person,ye2015coupled,zhong2017re,bai2019re}. Specifically, in  \cite{garcia2015person}, an unsupervised re-ranking model is learned by jointly considering the content and context information in the ranking list. And in \cite{ye2015coupled}, the shared nearest neighbors of different baseline methods are exploited to re-rank the task of interest. After that, Zhong \etal \cite{zhong2017re} further proposed a $k$-reciprocal encoding method and reported superior performance. Recently, by fusing the results of several different models, an effective re-ranking solution is further developed in \cite{bai2019re}. 
While most re-ranking methods start from distance space to improve the performance, our IIA works from the feature space by a weighted and iterative design and keeps updating the impression of a person of interest. In the following experiments, we will show that the proposed simple IIA model can be as effective as these advanced re-ranking methods. And by leveraging stronger representation and similarity metrics, IIA can further improve the re-ID performance.

For query expansion, it denotes the process of reformulating a given query and is widely used in image retrieval \cite{salton1990improving,chum2007total,arandjelovic2012three}. In particular, by concatenating the averaged top-k candidate galleries, a new query will be produced in the well-known average query expansion (AQE) method \cite{chum2007total}. And in  \cite{arandjelovic2012three}, AQE is further improved by  discriminative query expansion (DQE). Rather than using a simple averaging, a linear SVM is leveraged in DQE to obtain a new query and the distance from the decision boundary is employed for the initial ranking list revision. The proposed IIA generally shares the same spirit with \cite{chum2007total, arandjelovic2012three}, but involves a more advanced weighting mechanism for updating query and formulates this process in an iterative manner.

\subsection{Iterative updating}
The idea of updating an impression of a person iteratively is natural and intuitive. Technically, iterative updating has been well studied in the field of optimization  \cite{green1984iteratively,xin2016maximal,daubechies2010iteratively}. Our specific implementation is also highly motivated from the recent attention models originated from natural language processing \cite{vaswani2017attention,devlin2018bert} and memory networks  \cite{weston2014memory,sukhbaatar2015end}. In natural language processing, attention can be applied to different positions in a sentence to formulate an overall understanding of the sentence in a step-by-step manner. And in memory networks, such as \cite{weston2014memory} and the neural Turing machines \cite{graves2014neural}, memory represented as vectors can be erased, updated and read with mathematical tools. Our work, though not seemingly related, shares similar conceptual intuition (e.g. impression as to memory) and mathematical formulation (e.g. attention by similarity) with these interesting works.

\begin{figure*}[t]
    \centering
    \includegraphics[scale=0.43]{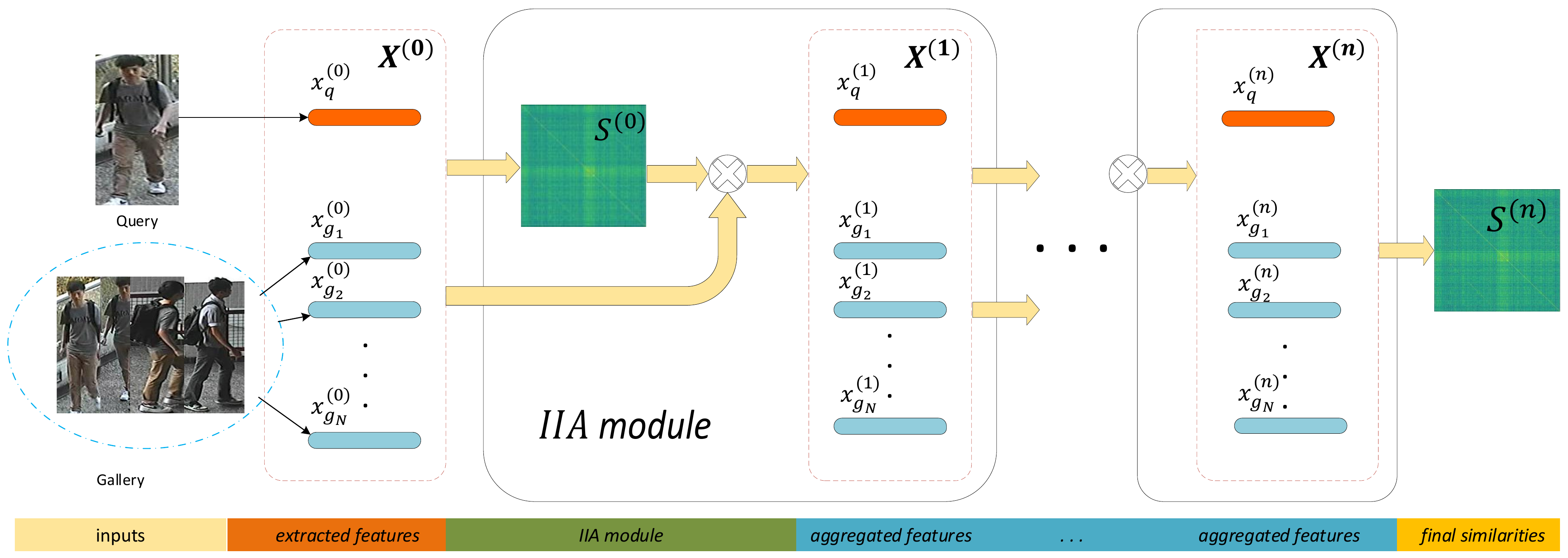}
    \caption{Illustration of the iterative impression aggregation pipeline. From left to right, we start from the input query and gallery images, we extract a baseline feature/representation, and iteratively apply the IIA module, where features from the last iteration will be updated by exploring similarities among images. After several iterations, the final impressions are formed and we use them to compute a final similarity for retrieval.}
    \label{fig:framework}
\end{figure*}

\section{Iterative Impression Aggregation}
\label{sec:method}

In human visual system, the impression  about  one  person keeps updating when we see more aspects of him/her, thus making the representation more and more robust. Since comprehensive and robust person representation is also required in person re-ID, we will formulate a model and put such an intuition computationally into practice in the the re-ID system  in this section.  We start by formulating a basic iterative attentional aggregation model in the settings of person re-ID in Section \ref{ssec:iaa} and discuss the extension to updating the representations of gallery candidates in Section \ref{ssec:giu}.Then we proceed to introduce more advanced similarity metrics to further strengthen the model in Section \ref{ssec:si} and provide computational complexity analysis to highlight the simplicity and efficiency of the proposed IIA in Section \ref{ssec:cc}.

Let us first consider a typical re-ID model, in which $\mathcal{G}$ is a gallery composed of $N$ images, denoted as $\{g_i\}_{i=1}^N$. Given a query image $q$, the task is to retrieve all candidates in $\mathcal{G}$ that is of the same identity as $q$. Different from the works about developing good person representations and similarity metrics from training data, the focus of our method is designing a strategy to better exploit information from the gallery dataset in an orthogonal way. Without loss of generality, we denote $\boldsymbol{x}_q$ and $\boldsymbol{x}_{g_i}$ as the developed/learned representations of the query and gallery images respectively and define $s_{qi}=s(q, g_i)$, $s_{ij}=s(g_i, g_j)$ as the metric developed for computing similarity (or distance) between the corresponding image pairs.

\subsection{Iterative Attentional Aggregation}
\label{ssec:iaa}

In order to iteratively aggregate new information from more aspects, we propose the following key formulation.
\begin{equation}
\label{eq:general}
    \boldsymbol{x}_q^{t+1} = m ( \boldsymbol{x}_q^t,  {f(\{s_{q1}, \boldsymbol{x}_{g_1}^t}\},..., \{s_{q_N}, \boldsymbol{x}_{g_N}^t\}))
\end{equation}
where $t$ is the iteration index. Note that, here we have two functions $m(\cdot)$ and $f(\cdot)$, which define a merging function and an attention function respectively. Two high level insights are under this general formulation: 1) We would like to update the representation of the query after we observe the gallery images and we would like to keep updating this representation iteratively. 2) During each iteration, the attention we paid to each gallery image is related to the similarity/distance between this image and the query image. By default, we use the following  simple function formulation.

\begin{equation}
\label{eq:merge}
    m(\boldsymbol{a}, \boldsymbol{b}) = \alpha \boldsymbol{a} + (1-\alpha) \boldsymbol{b}
\end{equation}
\begin{equation}
    \label{eq:att}
    f(s, x) = \sum_{g_i\in\mathcal{N}_{q}} {softmax}(s_{qi}) \boldsymbol{x}_{g_i} = \sum_{g_i\in\mathcal{N}_{q}} \frac{e^{s_{qi}/\tau}  \boldsymbol{x}_{g_i} }{\sum_{g_j\in\mathcal{N}_{q}}{e^{s_{qj}/\tau}}}
\end{equation}
where $m(\cdot)$ is a linear merging function that updates the impression with new information via a forgetting parameter $\alpha$. Intuitively, $\alpha$ can be used to control the updating intensity. $f(\cdot)$ is a straightforward soft attentional aggregation function defined upon $q$'s  top-k nearest neighbors $\mathcal{N}_{q}$ to avoid involving noisy samples. $\tau$ is the temperature factor to control the smoothness of the updating weight. Here we design $m(\cdot)$ and $f(\cdot)$ to be as simple and automatic as possible, which only involve three tuning parameters. However, more advanced merging and attention functions may be beneficial for larger performance gain. For example, LSTM cells or memory networks could be leveraged to automatically learn $m, f$. Due to the iterative nature, we can obtain a list of historical representations and similarities, it is also possible to apply all or part of the historical representations (or similarities) to achieve an ensemble when computing the final retrieval. Note that, the proposed IIA has one  good property, i.e., so long as $m(\cdot)$ and $f(\cdot)$ are differential with respect to $\boldsymbol{x}$, we can back-propagate information and further fine-tune the original representation $\boldsymbol{x}$. In this paper, we adopt the above light-weight design to prove the concept of IIA by default. This also allows us to put more efforts into equipping IIA with many different feature representations $\boldsymbol{x}$ and distance metrics $\bm{M}$.

In real application systems, if the number of candidates in the gallery is very large, we can simply apply IIA to smaller sampled subsets of the whole gallery. This can be sub-optimal, but could still potentially improve the performance with negligible computation overhead.

\begin{algorithm}[ht]
  \caption{IIA algorithm.}
  {
    Input: $\alpha, K, n, \bm{D}^{0}=[\boldsymbol{x}_q, \boldsymbol{x}_{g_1},..., \boldsymbol{x}_{g_N}] \in \mathbb{R}^{d\times(N+1)}$ \;
    Calculate initial similarity matrix: $\bm{S}^{0}\in \mathbb{R}^{(N+1)\times(N+1)}$ \;
    \For{t=1:n}{
      Keep top-K elements of $\bm{S}^{t-1}$ for each row: $\bm{S}^{t-1}_{topk}=topK(\bm{S}^{t-1})$\;
      Normalize  $\bm{S}^{t-1}_{topk}$ according to \ref{eq:att}\;
      Calculate Update matrix: $\bm{U}^{t}=\alpha \bm{I} + (1-\alpha)\bm{S}^{t-1}_{topk}$\;
      Update $\bm{D}$: $\bm{D}^{t}=\bm{D}^{t-1}\bm{U}^{t}$\;
      Calculate similarity matrix:$\bm{S}^{t}\in \mathbb{R}^{(N+1)\times(N+1)}$ \;
     }
    \Return{$\bm{S}^{n}$, $\bm{D}^{n}$.}
    \label{alg:alg}
    }
\end{algorithm}

\subsection{Gallery Impression Updates}
\label{ssec:giu}

In Eq \eqref{eq:general}, we only update the impression from the perspective of the query image. In this section, we will try to update the impression/representation of each gallery image given the rest of the gallery images and the query image in a similar manner. Although it is theoretically difficult to prove whether updating both query and gallery impressions enjoys a global optimal solution, we empirically observe that this choice often achieves better performance probably because more information is exchanged. 

To update both the query impression and the gallery impressions, we mathematically formulate a matrix
$ \bm{D} = [\boldsymbol{x}_q, \boldsymbol{x}_{g_1},..., \boldsymbol{x}_{g_N}] \in \mathbb{R}^{d\times(N+1)}$ ($d$ is the feature dimension) and keep track of a similarity matrix $\bm{S} \in \mathbb{R}^{(N+1)\times(N+1)}$. After we have computed the similarity matrix, we can set the diagonal elements of $\bm{S}$ to $\alpha$ and multiply $(1-\alpha)$ to each of the non-diagonal element to achieve an update matrix $\bm{U}$. Then the representations updates of both the query and gallery images can be obtained simultaneously via matrix multiplication $\bm{D}^{t+1} = \bm{D}^{t}\bm{U}$, which could be easily computed. We provide the complete process of IIA in Algorithm \ref{alg:alg} for better reproduction of this work.

\subsection{Similarity Incorporation and Enhancements}
\label{ssec:si}

Once we have the initial representations, simple Euclidean or cosine distances can be directly used as the similarity measurement $s$. In the later experiments, the effectiveness of such simple metrics will be demonstrated, showcasing that the idea of iterative impression update matters.

Of course, any other forms of  metrics could also be applied directly by leveraging better metric learning strategies from the re-ID literature. Note that, both re-ranking methods and our IIA try to exploit more information from the gallery dataset. But re-ranking methods work in the space of similarity metrics (thereafter ranking orders), while ours works in the space of representations. And because the proposed formulation also involves the similarity metrics to compute attention, we can potentially further benefit from incorporating other re-ranking methods to compute re-ranked similarities. Specifically, the idea can be achieved by first computing the initial similarity matrix $\bm{S}_{init}$ then applying any re-ranking methods to obtain an updated similarity matrix $\bm{S}_{new}$. Finally, $\bm{S}_{new}$ can be applied just as $\bm{S}$ shown in Algorithm \ref{alg:alg}.

\subsection{Computational Complexity}
\label{ssec:cc}

The main computational cost of the proposed IIA comes from pairwise distance/similarity computing and iterative feature updating. Suppose the size of the gallery set is $N$, the feature dimension is $d$, $K$ is the number of closest candidates for aggregation and $n$ is the number of iterations. For one query image, 
the computation complexity required for pairwise distance and the aggregation process are $O(Nnd)$ and $O(NnK+nKd))$, respectively. For updating galleries, the total computations are $O(N^2nd)+O(N^2nK+NnKd)$. In practice, gallery updates can be obtained offline, thus for a query $q$, the actual online computation is $O(Nnd+NnK+nKd)$. As a comparison, the computational complexity of state-of-the-art re-ranking method (RR) \cite{zhong2017re} is $O(N^2d)+O(N^2log(N))$ for offline pairwise distance computing and online sorting for galleries and $O(Nd)+O(Nlog(N)K)$ for online re-ranking. Table \ref{tab:comp} compares the detailed computational complexity. As we empirically observe, the optimal or necessary $n$ and $K$ are often far less than $N$, which bring the speed advantage for IIA, especially during the actual online phase. 
Table \ref{tab:rt} shows some typical run-time comparison on three different datasets, where we can observe that IIA runs practically 4$\sim$6 times faster than RR \cite{zhong2017re}. To further just the runtime efficiency on larger dataset, we also conduct experiments on the MSMT17 dataset (0.1M images), and our IIA (398.39s) is still more than 4 times faster than RR (1638.99s).

\begin{table}[t]
    \centering
    \caption{Computation complexity comparison between IIA and re-ranking method RR  \cite{zhong2017re}}
    \begin{tabular}{c|c|c}
        \hline
         method & offline & online \\
         \hline
         RR & \scriptsize{$O(N^2log(N)$+$N^2d)$} & \scriptsize{$O(Nlog(N)K$+$Nd)$} \\
         \hline
         IIA$_{bas}$ & \scriptsize{$O(N^2nd$+$N^2nK$+$NnKd)$} & \scriptsize{$O(Nnd$+$NnK$+$nKd)$} \\
         \hline
    \end{tabular}
    \label{tab:comp}
\end{table}

\begin{table}[ht]
    \centering
    \caption{Total test run-time on four standard datasets, we run each method ten times with settings of MGN in Table \ref{tab:duke}, \ref{tab:market} and \ref{tab:cuhk}.}
    \begin{tabular}{c|c|c}
        \hline
         dataset & RR& IIA$_{bas}$ \\
         \hline
        Market1501 & $71.49 \pm 0.30$s & $12.79 \pm 0.01$s \\
        DukeMTMC & $73.54 \pm 0.71$s & $14.77 \pm 0.03$s \\
        CUHK03-NP-lab & $14.16 \pm 0.42$s & $3.11 \pm 0.00$s \\
        CUHK03-NP-det & $15,26 \pm 0.05$s & $2.83 \pm 0.00$s \\
        \hline
    \end{tabular}
    \label{tab:rt}
\end{table}

\section{Experiments}
\label{sec:expr}

We conduct extensive experiments on many popular datasets including CUHK03, Market-1501 and DukeMTMC. These datasets have been widely used as standard benchmarks to evaluate the performance of person re-ID methods.
We start by introducing the basics of these datasets and the corresponding experimental setting in Section \ref{ssec:dap}. Then we demonstrate the improvement of IIA over competitive baseline models and comparison with alternative post-processing methods in Section \ref{ssec:iobm}. Then we provide detailed ablation study of our proposed framework in Section \ref{ssec:ablation_study}. After that, we perform hyper parameters analysis in the experiments setting in Section~\ref{ssec:hyper-parameter analysis}. Then we demonstrate the effectiveness of our methods by comparing with state-of-the-art methods in Section~\ref{ssec:cts}, Finally, we discuss the limitation and failure cases of the proposed methods in Section~\ref{ssec:limitation_and_failurecases}.

\subsection{Datasets and Experiment Protocols}
\label{ssec:dap}

\subsubsection{Datasets}
\label{ssec:datasets}

\begin{table}[ht]
    \centering
    \caption{The detailed statistics of the datasets used in our experiments, including the number of IDs (\#ID), the number of boxes (\#box), the number of cameras (\#cam) and the number of boxes for per valid gallery identity (\#box/g\_id).}
    \begin{tabular}{c|c|c|c|c|c}
    \hline
        Datasets & \#ID & \#box &  \#box/ID  & \#cam & \#box/g\_id \\
        \hline
        DukeMTMC & $1,404$ & $36,411$ & $25.9$ & $8$ & $17.5$ \\
        Market-1501 & $1,501$ & $32,668$ & $21.8$ & $6$ & $20.2$ \\
        CUHK03 & $1,467$ & $14,096$ & $9.6$ & $2$ & $7.6$ \\
        \hline
    \end{tabular}
    \label{tab:dataset}
\end{table}

\textbf{DukeMTMC} \cite{zheng2017unlabeled} is a recent large scale benchmark. It contains 36,411 images of 1,404 identities. Each identity is captured from 8 cameras. On average, each identity has about 26 images. The dataset is split into two parts: 16,522 images with 702 identities for training and 19,889 images with 702 identities for testing. In the testing phase, 2,228 images with 702 identities are used as the query set. Following the literature, we report the single-query evaluation results for this dataset.

\textbf{Market-1501} \cite{zheng2015scalable}, a standard large scale image-based re-ID dataset, contains 32,668 labeled bounding boxes of 1,501 identities captured from 6 different viewpoints. These bounding boxes are obtained by Deformable Parts Model (DPM)  \cite{felzenszwalb2008discriminatively}.
On average, each identity has about 20 images. The dataset is split into two parts: 12,936 images with 751 identities for training and 19,732 images with 750 identities for testing. In the testing phase, 3,368 hand-drawn images with 750 identities are used as the query set to identify the correct identities on the testing set. Following the literature, we report the single-query evaluation results for this dataset.

\textbf{CUHK03} \cite{li2014deepreid} a classic re-ID dataset, contains 14,096 images of 1,467 identities. Each identity is captured from two cameras on the campus of CUHK. On average, each identity has about 5 images in each camera. This dataset provides both manually labeled bounding boxes and DPM-detected bounding boxes. We report results in both the ‘labeled’ (CUHK03-lab) and ‘detected’ (CUHK03-det) settings. The dataset is split into two parts: 7,365 images with 767 identities for training and 6,732 images with 700 identities for testing. In the testing phase, 1,400 images with 700 identities are used as the query set. As we use the new protocols proposed by  \cite{zhong2017re}, we refer this dataset to CUHK03-NP in the rest of this paper.

\subsubsection{Evaluation Protocols}
\label{ssec:proc}
In all the experiments, we follow the standard evaluation metrics: mean Average Precision (mAP) and the Cumulated Matching Characteristics top-1 (cmc1) metric. On Market-1501 and DukeMTMC, our evaluation follows the official standard settings, and our evaluation on CUHK03 is performed by following the same setups as \cite{zhong2017re} for fair comparison.

\textbf{Baseline models.} We explore four representative baseline models: Trip  \cite{hermans2017defense}, where representative triplet loss is applied and PCB \cite{sun2018PCB} and PAB \cite{suh2018part}, where state-of-the-art part-level features are used and MGN  \cite{wang2018learning}, where latest state-of-the-art performance is reported. To demonstrate the effectiveness of our method, we combine two different versions of IIA with these baselines: a basic simple  IIA$_{bas}$ where we directly applied cosine similarity in the representation space, and a more advanced IIA$_{adv}$ where we utilize the re-ranked similarity from \cite{zhong2017re} to exploit auxiliary information as mentioned above.

\subsection{Improvements over Baseline Models}
\label{ssec:iobm}
In this section, on the one hand, we want to show the relative performance gain brought by IIA upon the baseline models. On the other hand, we compare the proposed IIA with two other gallery information exploitation methods: the state-of-the-art re-ranking method RR \cite{zhong2017re} and an effective query expansion method (AQE) \cite{chum2007total}. For extensive comparison, we integrate IIA, RR and AQE with the above mentioned Trip, PCB, PAB and MGN respectively. Table \ref{tab:duke},\ref{tab:market} and \ref{tab:cuhk} show the detailed comparison results on the Market-1501, DukeMTMC, CUHK03-NP detected and labeled benchmarks. The performance of baseline models is obtained with released models or source code from the original papers. And we report the results of RR and AQE methods by using default parameters mentioned in their papers. For our method, we set the hyper-parameters by tuning on the validation dataset. More discussions about these hyper-parameters will also be provided in the following analysis part.

\begin{table}[t]
    \centering
    \caption{Performance comparison on DukeMTMC, the results from our models are achieved with hyper-parameter: $\alpha$=0.82, $K$=13, $n$=7, where $\alpha$ is the forgetting parameter, $K$ is the number of closest candidates for aggregation and $n$ is the final number of iterations.}
    \begin{tabular}{ p{28mm} | >{\centering\arraybackslash}p{16mm} | >{\centering\arraybackslash}p{16mm}}
        \hline
        Method &  mAP &  cmc1 \\
        \hline
        
        \hline
        Trip \cite{hermans2017defense} & $63.92$ & $79.53$ \\
        Trip+AQE \cite{chum2007total} & $74.87$ & $82.81$ \\
        Trip+RR \cite{zhong2017re} & $81.30$ & $84.47$ \\
        Trip+RR+AQE & $76.22$ & $84.56$ \\
        Trip+IIA$_{bas}$ & $78.45$ & $83.84$ \\
        Trip+IIA$_{adv}$ & $\mathbf{82.14}$ & $\mathbf{85.10}$ \\
        \hline
        PCB \cite{sun2018PCB} & $69.94$ & $84.47$ \\
        PCB+AQE \cite{chum2007total} & $80.20$ & $87.21$ \\
        PCB+RR \cite{zhong2017re} & $84.59$ & $88.82$ \\
        PCB+RR+AQE & $80.77$ & $87.97$ \\
        PCB+IIA$_{bas}$ & $83.21$ & $88.02$ \\
        PCB+IIA$_{adv}$ & $\mathbf{85.20}$ & $\mathbf{88.82}$ \\
        \hline
        PAB \cite{suh2018part} & $68.43$ & $84.20$ \\
        PAB+AQE \cite{chum2007total} & $79.04$ & $87.39$ \\
        PAB+RR \cite{zhong2017re} & $84.21$ & $88.42$ \\
        PAB+RR+AQE & $80.11$ & $88.69$ \\
        PAB+IIA$_{bas}$ & $82.81$ & $88.62$ \\
        PAB+IIA$_{adv}$ & $\mathbf{85.27}$ & $\mathbf{88.78}$ \\
        \hline
        MGN \cite{wang2018learning} & $76.88$ & $88.33$ \\
        MGN+AQE \cite{chum2007total} & $86.70$ & $91.02$ \\
        MGN+RR \cite{zhong2017re} & $90.04$ & $92.06$ \\
        MGN+RR+AQE & $87.07$ & $91.47$ \\
        MGN+IIA$_{bas}$ & $89.40$ & $91.11$ \\
        MGN+IIA$_{adv}$ & $\mathbf{90.71}$ & $\mathbf{92.24}$ \\
        \hline
    \end{tabular}
    
    \label{tab:duke}
\end{table}

\begin{table}[t]
    \centering
    \caption{Performance comparison on Market-1501, the results from our models are achieved with hyper-parameter: $\alpha$=0.82, $K$=11, $n$=6, where $\alpha$ is the forgetting parameter, $K$ is the number of closest candidates for aggregation and $n$ is the final number of iterations.}
    \begin{tabular}{ p{28mm} | >{\centering\arraybackslash}p{16mm} | >{\centering\arraybackslash}p{16mm}}
        \hline
        Method & mAP & cmc1 \\
        \hline
        
        \hline
        Trip \cite{hermans2017defense} & $74.85$ & $88.33$ \\
        Trip+AQE \cite{chum2007total} & $83.09$ & $90.02$ \\
        Trip+RR \cite{zhong2017re} & $87.68$ & $91.36$ \\
        Trip+RR+AQE & $84.49$ & $\mathbf{91.48}$ \\
        Trip+IIA$_{bas}$ & $85.56$ & $90.66$ \\
        Trip+IIA$_{adv}$ & $\mathbf{87.89}$ & $91.24$ \\
        \hline
        PCB \cite{sun2018PCB} & $78.54$ & $92.87$ \\
        PCB+AQE \cite{chum2007total} & $87.00$ & $93.53$ \\
        PCB+RR \cite{zhong2017re} & $90.16$ & $\mathbf{94.06}$ \\
        PCB+RR+AQE & $87.80$ & $93.88$ \\
        PCB+IIA$_{bas}$ & $89.02$ & $93.56$ \\
        PCB+IIA$_{adv}$ & $\mathbf{90.48}$ & $94.03$ \\
        \hline
        PAB \cite{suh2018part} & $79.35$ & $92.28$ \\
        PAB+AQE \cite{chum2007total} & $86.83$ & $92.73$ \\
        PAB+RR \cite{zhong2017re} & $90.10$ & $93.40$ \\
        PAB+RR+AQE & $87.77$ & $\mathbf{93.50}$ \\
        PAB+IIA$_{bas}$ & $88.59$ & $93.05$ \\
        PAB+IIA$_{adv}$ & $\mathbf{90.41}$ & $93.44$ \\
        \hline
        MGN \cite{wang2018learning} & $86.00$ & $95.19$ \\
        MGN+AQE \cite{chum2007total} & $92.82$ & $95.87$ \\
        MGN+RR \cite{zhong2017re} & $94.11$ & $95.64$ \\
        MGN+RR+AQE & $93.08$ & $\mathbf{95.93}$ \\
        MGN+IIA$_{bas}$ & $93.60$ & $95.84$ \\
        MGN+IIA$_{adv}$ & $\mathbf{94.50}$ & $95.69$ \\
        \hline
    \end{tabular}
    \label{tab:market}
\end{table}

It can be easily observed that both the simple and advanced IIA are able to boost the baseline models' performance by a large margin, especially on the evaluation metric of mAP. Specifically, our IIA$_{bas}$ has an improvement of $16.29\%\sim22.73\%$, $8.84\%\sim14.31\%$ and $26.98\%\sim33.29\%$ among different baseline models for DukeMTMC, Market-1501 and CUHK03-NP respectively. And for cmc1, the corresponding improvements are $3.15\%\sim5.42\%$, $0.68\%\sim2.64\%$ and $14.96\%\sim20.33\%$.

Compared to the classic query expansion solution AQE \cite{chum2007total}, our IIA consistently outperforms it, highlighting the advantage of our elaborated attentional design and its iterative update mechanism. And compared to state-of-the-art re-ranking method RR \cite{zhong2017re}, IIA can achieve better performance in most cases while ensuring much faster speed. We further try to combine RR and AQE, and results show that a better initial ranking is also beneficial to AQE. However, the improvement is minor and the final performance is still worse than our IIA$_{bas}$ with a certain margin. Moreover, we find RR+AQE is even weaker than RR on mAP, which indicates that a naive query expansion with averaged top-k galleries at one time potentially cannot fully utilize the useful extra information brought by a strong initial ranking.

When comparing the results of IIA$_{adv}$ with the simple IIA$_{bas}$, we observe that, though IIA$_{adv}$ incorporates re-ranking similarities, IIA$_{adv}$ does not always bring extra performance gain over IIA$_{bas}$. This showcases that IIA$_{bas}$ is already a very competitive (and faster) alternative. But on DukeMTMC and Market-1501 datasets where the galleries are large and more gallery information is able to be exploited, the enhanced similarity can bring extra performance gain.

\begin{figure}[t]
    \centering
    \includegraphics[width=3.0in]{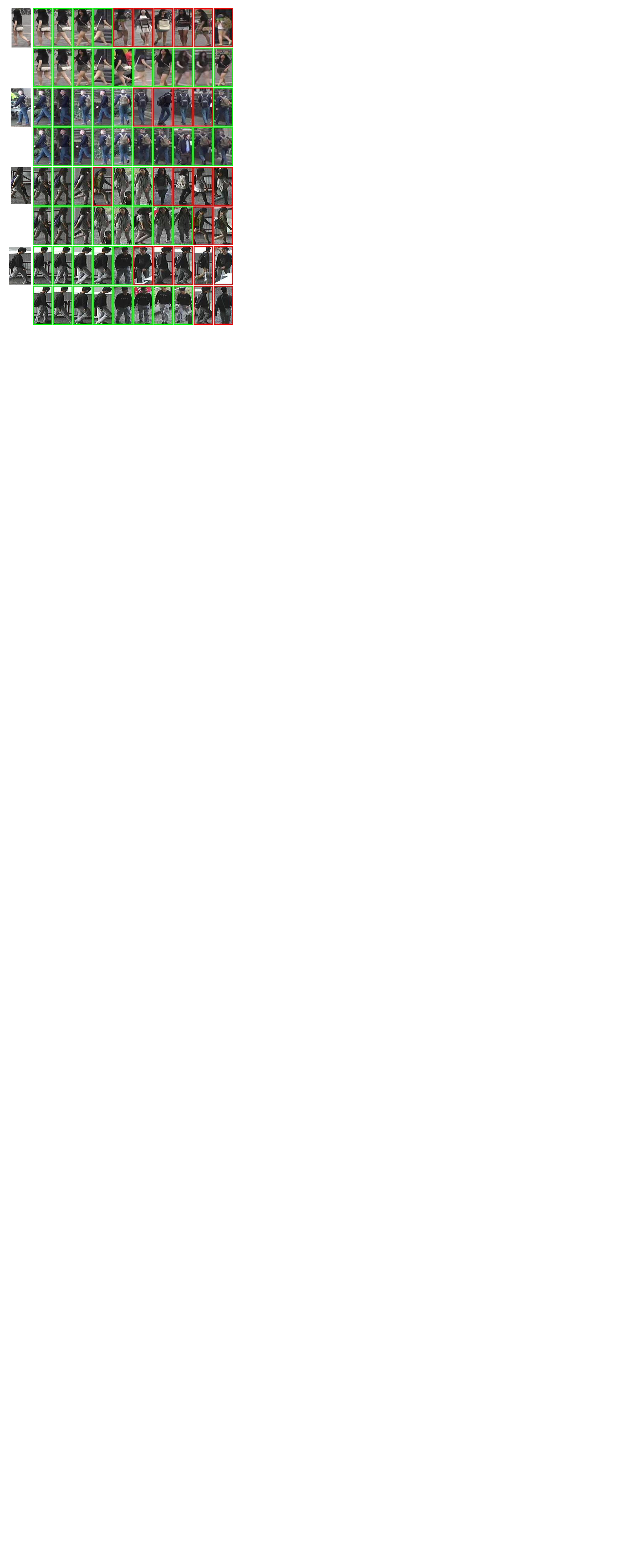}
    \caption{Examples of updated ranking from new impressions (from top to bottom these samples come from Market1501, DukeMTMC, CUHK03-NP-lab, CUHK03-NP-det respectively). For each query image (the left most ones), the first row shows the original ranking list produced by PCB, while the second row demonstrates the updated ranking results obtained by IIA$_{bas}$. Images surrounded by green box denotes its identity is the same with that of the query image and the red-boxed images are negative ones. }
    \label{fig:vis}
\end{figure}

In conclusion, the proposed IIA can consistently improve competitive baseline models and outperform both classic query expansion method and state-of-the-art re-ranking method in most scenarios. It demonstrates that IIA can effectively take advantage of the gallery information for person re-ID. And thanks to the simplicity of IIA$_{bas}$, it could be a light-weight off-the-shelf module for existing re-ID models to improve their performance.
In Fig. \ref{fig:vis}, we provide some representative examples to illustrate the updated ranking with new representations obtained by IIA. Obviously, the updated impressions/representations from IIA help to better rank the gallery candidates and move many true positives to the front.

\begin{table}[ht]
    \centering
    \caption{Performance comparison on CUHK03-NP dataset, the results from our models are achieved with hyper-parameter: $\alpha$=0.82, $K$=8, $n$=12, where $\alpha$ is the forgetting parameter, $K$ is the number of closest candidates for aggregation and $n$ is the final number of iterations.}
    \begin{tabular}{ p{22mm} | >{\centering\arraybackslash}p{10mm} | >{\centering\arraybackslash}p{10mm} | 
    >{\centering\arraybackslash}p{10mm} | >{\centering\arraybackslash}p{10mm} }
        \hline
        \multirow{2}{*}{Method} &
        \multicolumn{2}{ >{\centering\arraybackslash}p{20mm} | }{CUHK03-NP-lab} & \multicolumn{2}{ >{\centering\arraybackslash}p{20mm} }{CUHK03-NP-det} \\
        \cline{2-5} & mAP & cmc1 & mAP & cmc1 \\
        \hline
        
        \hline
        Trip \cite{hermans2017defense} & $57.81$ & $63.29$ & $54.85$ & $60.79$ \\
        Trip+AQE \cite{chum2007total} & $69.45$ & $69.93$ & $66.39$ & $67.71$ \\
        Trip+RR \cite{zhong2017re} & $73.23$ & $71.71$ & $70.16$ & $69.57$ \\
        Trip+RR+AQE & $71.83$ & $72.43$ & $69.09$ & $70.86$ \\
        Trip+IIA$_{bas}$ & $74.70$ & $73.50$ & $72.58$ & $72.07$ \\
        Trip+IIA$_{adv}$ & $\mathbf{75.53}$ & $\mathbf{73.86}$ & $\mathbf{72.64}$ & $\mathbf{72.10}$ \\
        \hline
        PCB \cite{sun2018PCB} & $61.57$ & $68.07$ & $57.53$ & $62.86$ \\
        PCB+AQE \cite{chum2007total} & $74.11$ & $74.86$ & $70.03$ & $71.79$ \\
        PCB+RR \cite{zhong2017re} & $76.65$ & $75.64$ & $74.11$ & $74.14$ \\
        PCB+RR+AQE & $76.43$ & $77.21$ & $73.55$ & $75.07$ \\
        PCB+IIA$_{bas}$ & $\mathbf{79.78}$ & $\mathbf{78.50}$ & $76.68$ & $\mathbf{75.64}$ \\
        PCB+IIA$_{adv}$ & $78.66$ & $76.93$ & $\mathbf{76.70}$ & $75.58$ \\
        \hline
        PAB \cite{suh2018part} & $56.19$ & $61.86$ & $55.20$ & $57.79$ \\
        PAB+AQE \cite{chum2007total} & $67.86$ & $69.07$ & $63.95$ & $64.57$ \\
        PAB+RR\cite{zhong2017re} & $72.12$ & $71.71$ & $69.11$ & $65.79$ \\
        PAB+RR+AQE & $70.54$ & $71.21$ & $66.21$ & $66.79$ \\
        PAB+IIA$_{bas}$ & $72.98$ & $\mathbf{71.86}$ & $70.93$ & $68.21$ \\
        PAB+IIA$_{adv}$ & $\mathbf{73.13}$ & $71.64$ & $\mathbf{71.39}$ & $\mathbf{68.93}$ \\
        \hline
        MGN \cite{wang2018learning} & $67.67$ & $71.93$ & $64.43$ & $69.71$ \\
        MGN+AQE \cite{chum2007total} & $80.29$ & $80.57$ & $77.32$ & $76.79$ \\
        MGN+RR \cite{zhong2017re} & $82.87$ & $82.07$ & $79.64$ & $78.29$ \\
        MGN+RR+AQE & $82.94$ & $82.86$ & $79.16$ & $79.14$ \\ 
        MGN+IIA$_{bas}$ & $\mathbf{85.93}$ & $\mathbf{84.93}$ & $82.72$ & $80.14$ \\
        MGN+IIA$_{adv}$ & $85.66$ & $84.36$ & $\mathbf{82.74}$ & $\mathbf{80.21}$ \\
        \hline
    \end{tabular}
    \label{tab:cuhk}
\end{table}

\begin{table}[ht]
    \centering
    \caption{Results for Combining Metric Learning and IIA$_{bas}$ on Market-1501 dataset.}
    \begin{tabular}{ p{31mm} | >{\centering\arraybackslash}p{16mm} | >{\centering\arraybackslash}p{16mm} }
        \hline
        Method & mAP & cmc1 \\
        \hline
        
        \hline
        LOMO \cite{liao2015person} & $8.01$ & $27.14$ \\
        LOMO+IIA$_{bas}$ & $15.54$ & $29.93$ \\
        LOMO+KISSME \cite{koestinger2012large} & $19.33$ & $42.19$ \\
        LOMO+XQDA \cite{liao2015person} & $22.03$ & $43.47$ \\
        LOMO+KISSME+IIA$_{bas}$ & $29.46$ & $47.54$ \\
        LOMO+XQDA+IIA$_{bas}$ & $\mathbf{30.80}$ & $\mathbf{47.54}$ \\
        \hline
        BOW \cite{zheng2015scalable} & $11.31$ & $31.77$ \\
        BOW+IIA$_{bas}$ & $16.59$ & $36.64$ \\
        BOW+KISSME \cite{koestinger2012large} & $16.94$ & $40.26$ \\
        BOW+XQDA \cite{liao2015person} & $11.99$ & $31.32$ \\
        BOW+KISSME+IIA$_{bas}$ & $\mathbf{21.79}$ & $\mathbf{41.78}$ \\
        BOW+XQDA+IIA$_{bas}$ & $21.20$ & $39.70$ \\
        \hline
        MGN \cite{wang2018learning} & $86.00$ & $95.19$ \\
        MGN+KISSME \cite{koestinger2012large} & $86.69$ & $94.98$ \\
        MGN+XQDA \cite{liao2015person} & $87.10$ & $94.60$ \\
        MGN+KISSME+IIA$_{bas}$ & $\mathbf{93.00}$ & $\mathbf{95.69}$ \\
        MGN+XQDA+IIA$_{bas}$ & $92.94$ & $95.19$ \\
        \hline
    \end{tabular}
    \label{tab:ml}
\end{table}

Another interesting investigation is to see how IIA collaborates with metric learning ideas, which are often utilized by person re-ID methods to boost their performance. Our work can also perform quite well with metric learning methods. To validate this, we conduct experiments based on methods including XQDA \cite{liao2015person} and KISSME \cite{koestinger2012large} with LOMO \cite{liao2015person}, BOW \cite{zheng2015scalable} and MGN \cite{wang2018learning} features on the Market-1501 dataset. Table \ref{tab:ml} shows that IIA can also improve the performance on naive hand-craft features. Even for well trained deep features (e.g. MGN), where the classification boundary in the training set is already very clear (accuracy$\approx$1) and additional metric learning is unable to bring obvious improvements, facilitated with IIA, we can make a considerable improvement over different features.
Therefore, IIA seems to be able to explore useful gallery information that could have been neglected by conventional designs including but may not limited to one-step query expansion, re-ranking and metric learning etc.

\subsection{Ablation Study}
\label{ssec:ablation_study}
In the proposed IIA, two important ingredients are iterative updating and query/gallery updating. To demonstrate their importance, we provide detailed analysis experiments and important visualizations to facilitate our understanding. 
\subsubsection{Effectiveness of Iterative updating}
\begin{figure}[ht]
    \centering
    \includegraphics[width=3.2in]{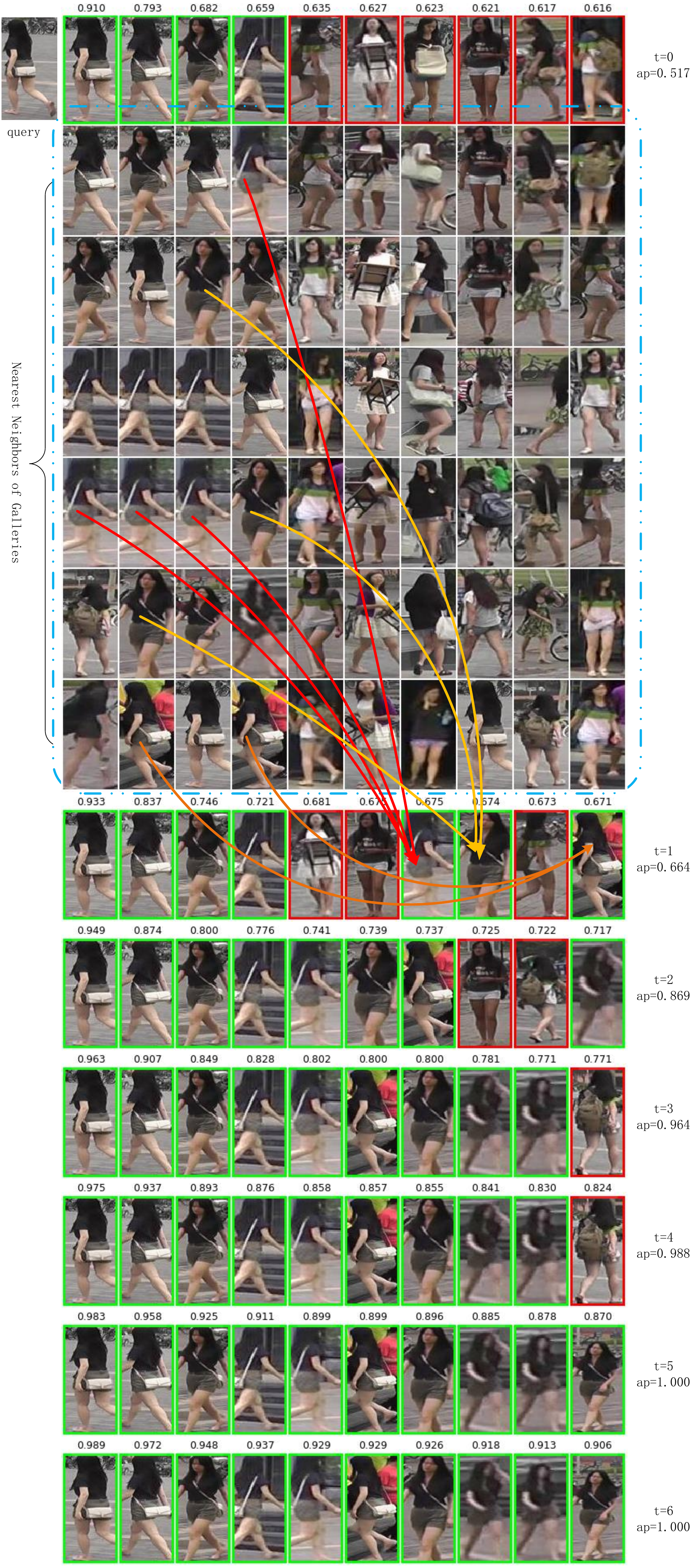}
    \caption{Visualization of the iterative updating. Images surrounded by green box denotes that its identity is the same as that of the query image and the red-boxed images are negative ones. The number above each image shows the similarity with query. Each column in the blue dashed box contains top-6 nearest neighbors of the corresponding gallery image in the first row (aka iteration 0) of the figure.}
    \label{fig:vis2}
\end{figure}

To illustrate the iterative updating mechanism, we demonstrate one typical example in Fig. \ref{fig:vis2}. The initial ranking ($t=0$) is given in the first row, in which only the top-4 galleries are correct. However, since the top-4 candidates have some neighbors in common (line marked as same color) from the rest of the gallery images, after updating the query representation with candidates' impressions, the similarities between query and these common galleries are enhanced. Therefore, when these missed common galleries (which are actually true positive candidates) were moved forward in the ranking list in the next iteration, and the average precision (AP) increases from 0.517 to 0.664 as well. As such procedure keeps iterating, more and more true galleries will be dug out with higher similarity. The top-10 list converges starting from $t=5$.

\subsubsection{Necessity of query and gallery updating}
\label{ssec:aa_nec}
When applying IIA, we empirically consider both query impression updating and gallery impression updating, both of which were testified as very useful in our experiments. In particular, in Table \ref{tab:iia-set}, the performance of three different settings is provided. When only query is updated (`q only'),  the original baselines can be significantly boosted, which shows that iterative query updating is already effective. When we update the gallery representation, we further improve the performance by a large margin. For gallery updating, `g online' means updating gallery representation at every iteration and meanwhile updating query with each iteration's updated gallery representation; while `g offline' means updating gallery $n$ iterations in advance and use the final iteration's gallery representation to update every iteration's query. Compared to `g online', `g offline' is much faster by reducing the updating frequency especially when the gallery set is very large. It can be seen that the offline version performs slightly worse than its online counterpart, but is still very competitive. In real-time systems, `g offline' should be a good choice which significantly reduces the computation time without dropping much performance. 

\begin{table}[h]
    \centering
    \caption{Results for different IIA updating settings for MGN features.}
    \begin{tabular}{ p{12.5mm} | >{\centering\arraybackslash}p{12.5mm} | >{\centering\arraybackslash}p{12.5mm} | >{\centering\arraybackslash}p{12.5mm} | >{\centering\arraybackslash}p{12.5mm} }
        \hline
        \multirow{2}{*}{Setting} & \multicolumn{2}{ >{\centering\arraybackslash}p{25mm} | }{DukeMTMC} & \multicolumn{2}{ >{\centering\arraybackslash}p{25mm} }{Market1501}  \\
        \cline{2-5} & mAP & cmc1 & mAP & cmc1 \\
        \hline
        
        \hline
        Original & $76.88$ & $88.33$ & $86.00$ & $95.19$ \\
        q only & $83.85$ & $\mathbf{91.83}$ & $89.36$ & $94.83$  \\
        g offline & $88.60$ & $90.80$ & $93.08$ & $95.40$ \\
        g online & $\mathbf{89.40}$ & $91.11$ & $\mathbf{93.60}$ & $\mathbf{96.84}$  \\ 
        \hline
        \hline
        \multirow{2}{*}{Setting} & \multicolumn{2}{ >{\centering\arraybackslash}p{25mm} | }{CUHK03-NP-lab} & \multicolumn{2}{ >{\centering\arraybackslash}p{25mm} }{CUHK03-NP-det} \\
        \cline{2-5} & mAP & cmc1 & mAP & cmc1\\
        \hline
        
        \hline
        Original & $67.67$ & $71.93$ & $64.43$ & $69.71$ \\
        q only & $71.55$ & $71.79$ & $69.08$ & $69.29$ \\
        g offline & $82.16$ & $82.07$ & $78.52$ & $78.14$ \\
        g online & $\mathbf{85.93}$ & $\mathbf{84.93}$ & $\mathbf{82.72}$ & $\mathbf{80.14}$ \\ 
        \hline
    \end{tabular}
    \label{tab:iia-set}
\end{table}

\subsection{Hyper Parameters Analysis}
\label{ssec:hyper-parameter analysis}
In our method, there are some key hyper-parameters that can influence the final performance. To study the effects of each hyper-parameter and get more insights and understanding of them, we provide detailed analysis in this regard. 

\subsubsection{Forgetting factor $\alpha$}
\begin{figure}[ht]
    \centering
    \includegraphics[width=1.62in]{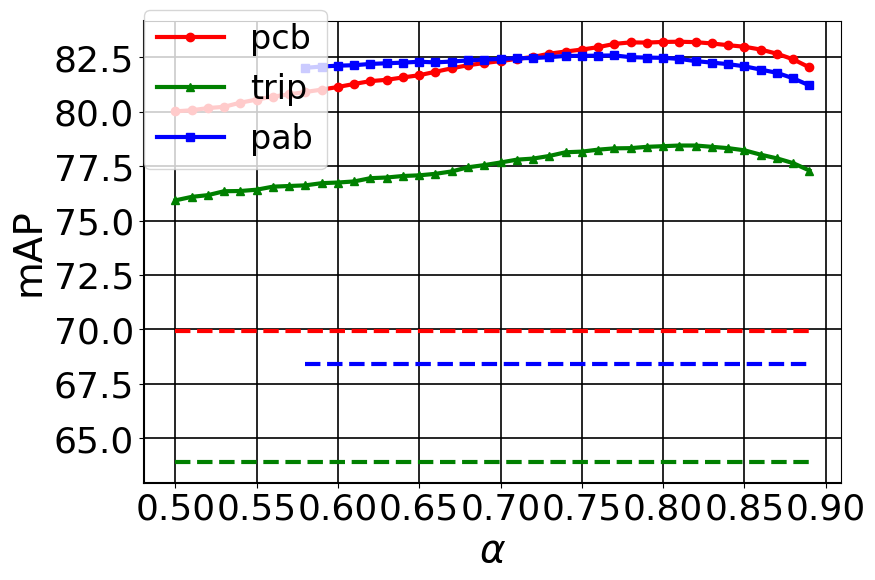}
    \includegraphics[width=1.62in]{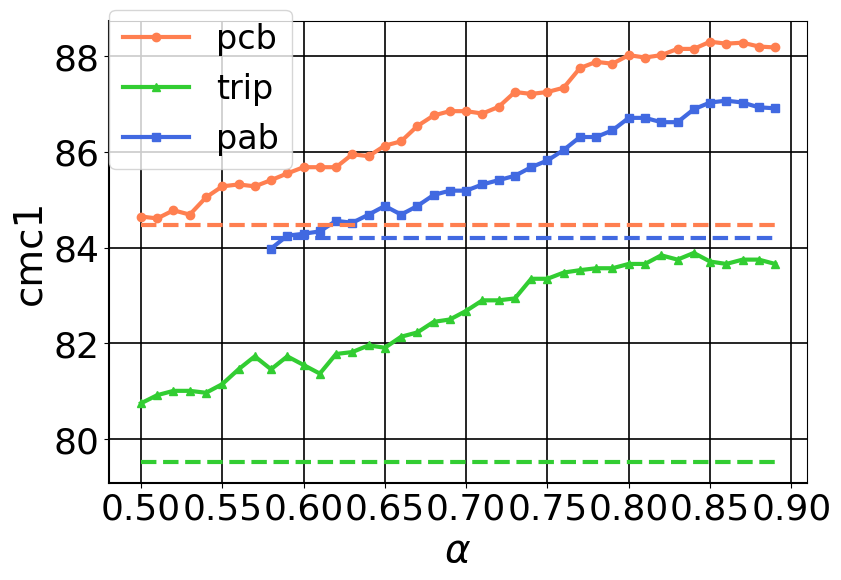}
    \includegraphics[width=1.62in]{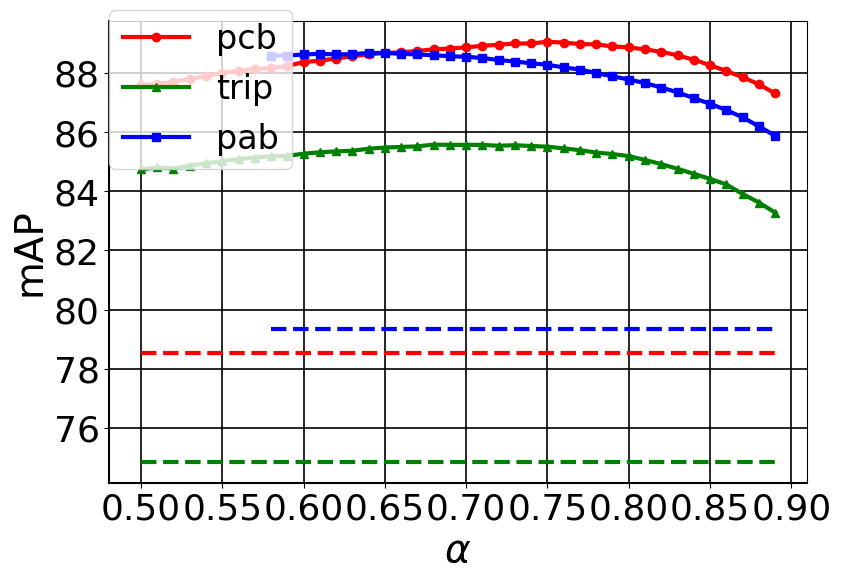}
    \includegraphics[width=1.62in]{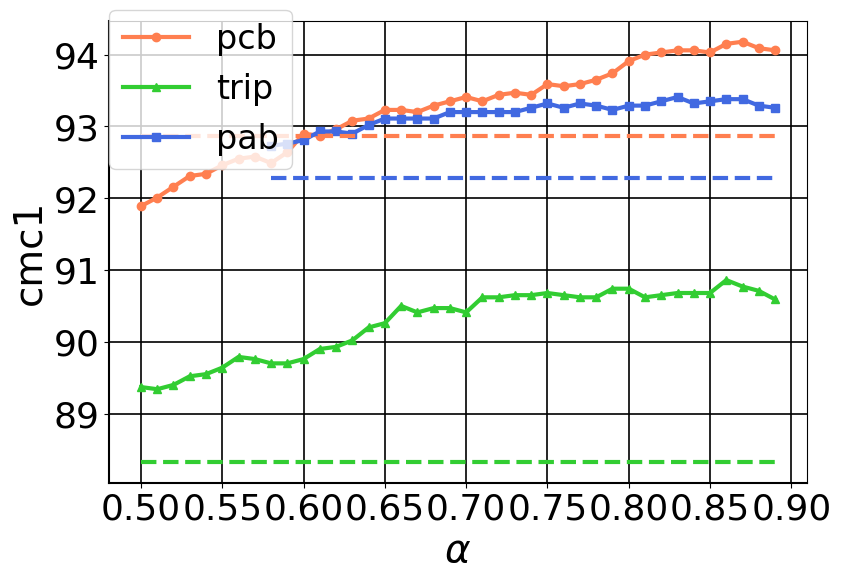}
    \includegraphics[width=1.62in]{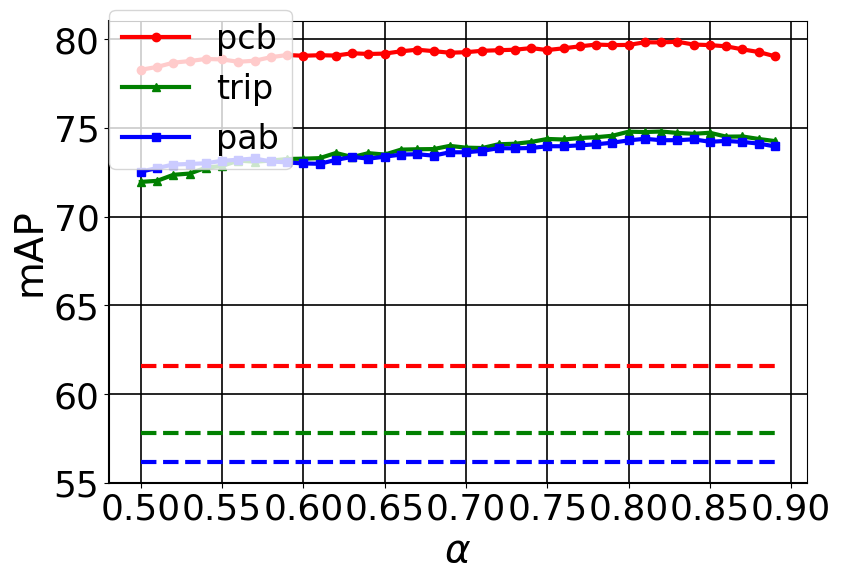}
    \includegraphics[width=1.62in]{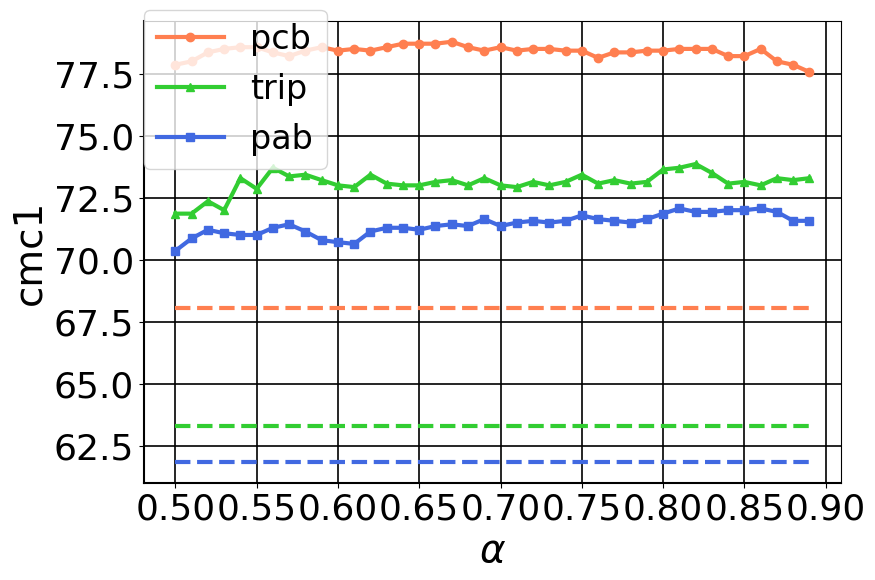}
    \includegraphics[width=1.62in]{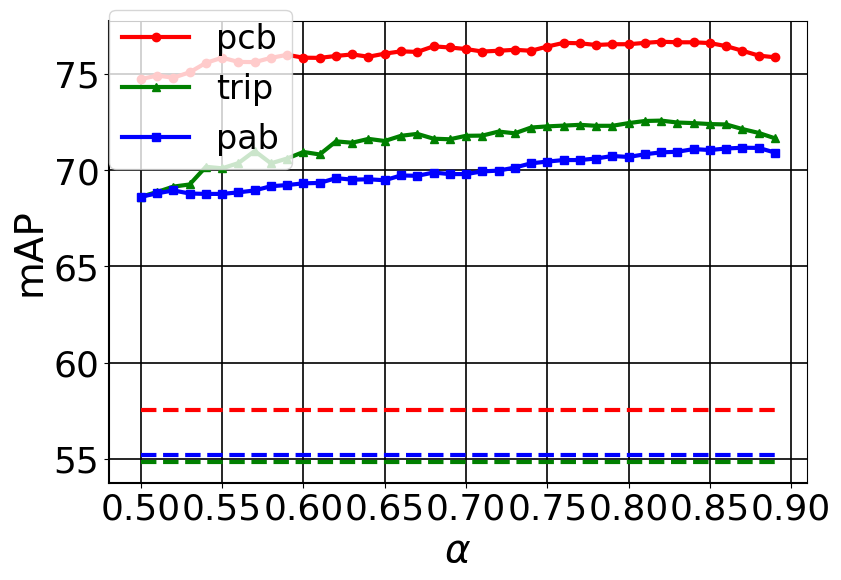}
    \includegraphics[width=1.62in]{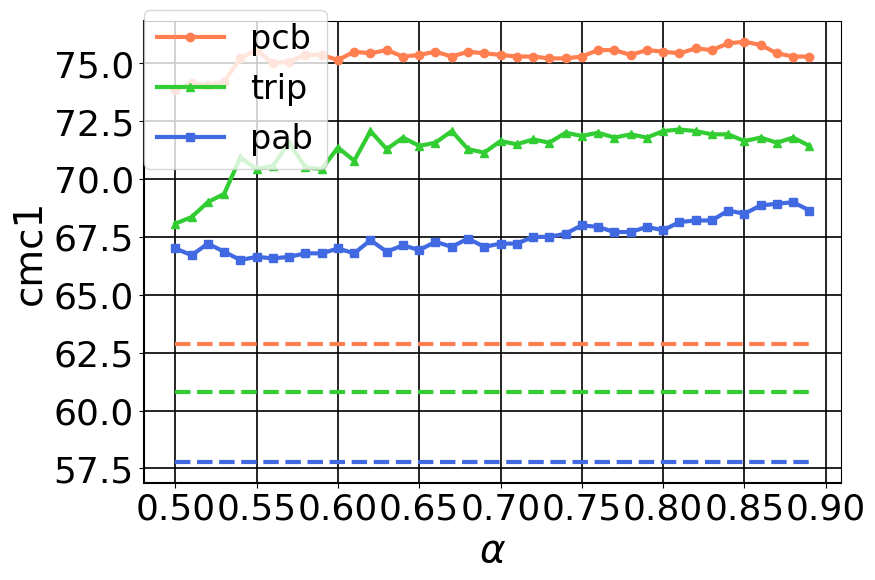}
    \caption{Performances (left: mAP; right cmc1) change with respect to $\alpha$ on DukeMTMC, Market-1501, CUHK03-NP-lab, CUHK03-NP-det (from top to bottom) validation set, we fix hyper-parameters $K$ and $n$ according to Table\ref{tab:duke}, \ref{tab:market} and \ref{tab:cuhk}. The dashed line is the baseline.}
    \label{fig:param_alpha}
\end{figure}

In Eq \ref{eq:merge}, $\alpha$ controls how much impression information should be updated at each iteration. We illustrate how the performance of both mAP and cmc1 change with respect to $\alpha$ in Fig. \ref{fig:param_alpha}. As is shown, our method consistently outperforms the baselines both on the mAP and cmc1 with a wide range of $\alpha$ for all the three datasets. Specifically, the mAP increases along as $\alpha$ increases, before it begins to decrease rapidly after $\alpha$ surpasses a threshold. As demonstrated from this figure, the optimal $\alpha$ for mAP is roughly around 0.8. Cmc1 shares similar behavior to mAP, but the optimal $\alpha$ is larger (around 0.9). To our knowledge, since cmc1 only considers whether the largest gallery candidate is consistent with the query, it tends to have very low tolerance for involving wrong information. Therefore, it prefers a more conservative updating mechanism. In contrast, since mAP considers both precision and recall, it will have an improvement as long as the updated information has more constructive effect than destructive. So mAP will prefer a lower $\alpha$ to let more information in from galleries. In this paper, we prefer to select optimal positions according to mAP (since it considers both recall and precision) and choose $\alpha=0.82$ by default. Furthermore, we observe that the optimal $\alpha$ is empirically stable across datasets, indicating that the proposed IIA is robust to $\alpha$ to some extent. In a nutshell, we do recommend users to set $\alpha$ around 0.8 when come across unknown datasets.

\subsubsection{Number of iteration $n$}
\begin{figure}[ht]
    \centering
    \includegraphics[width=1.62in]{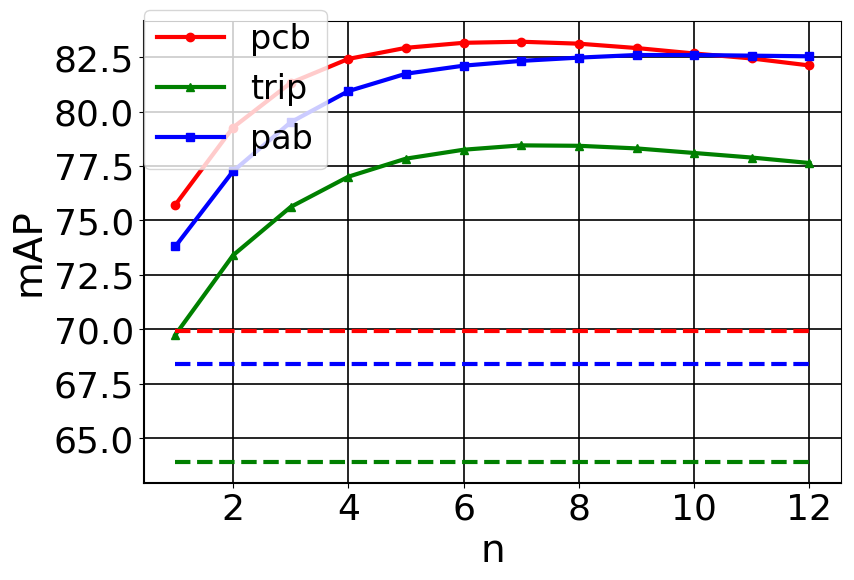}
    \includegraphics[width=1.62in]{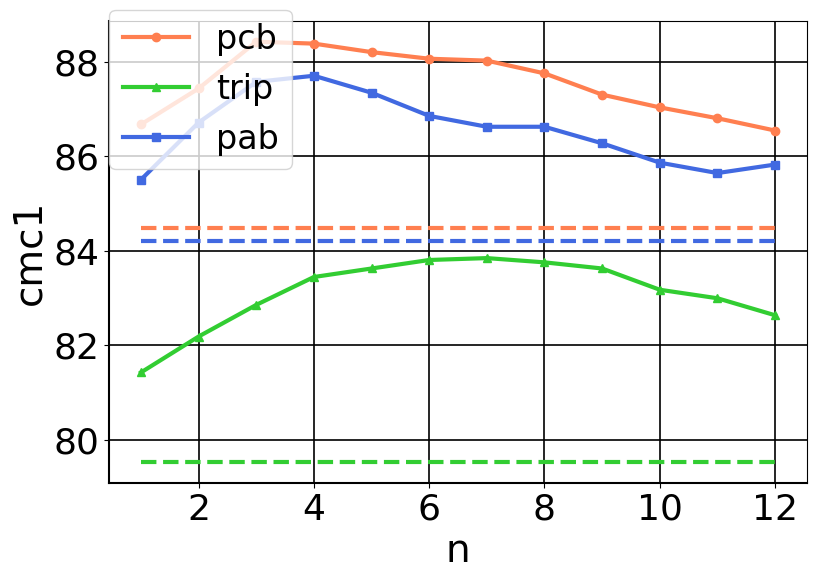}
    \includegraphics[width=1.62in]{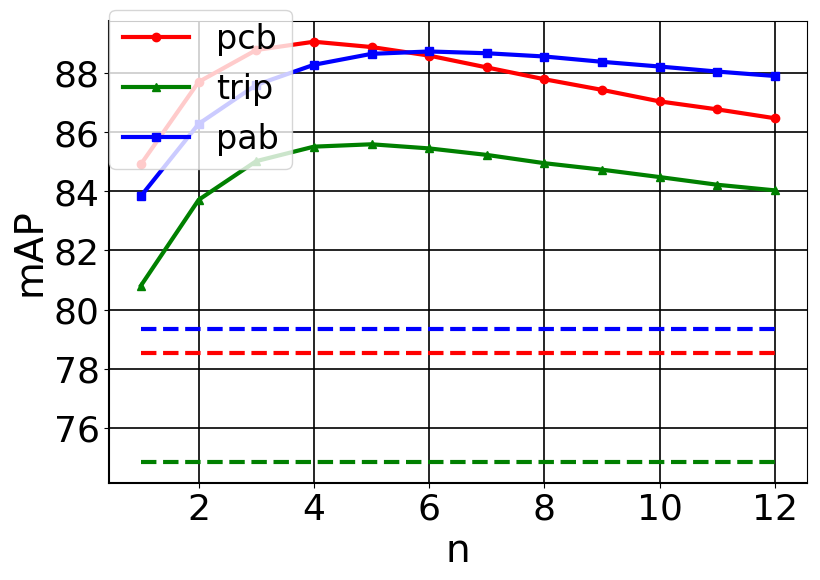}
    \includegraphics[width=1.62in]{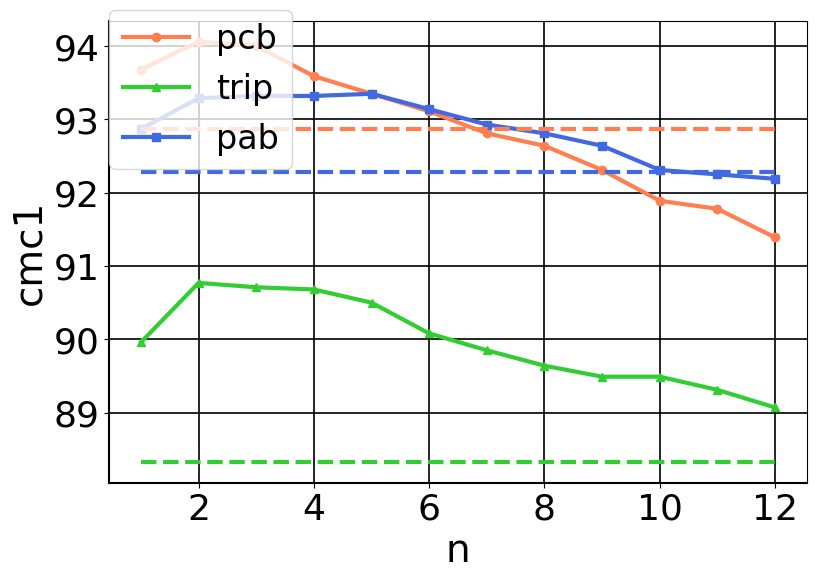}
    \includegraphics[width=1.62in]{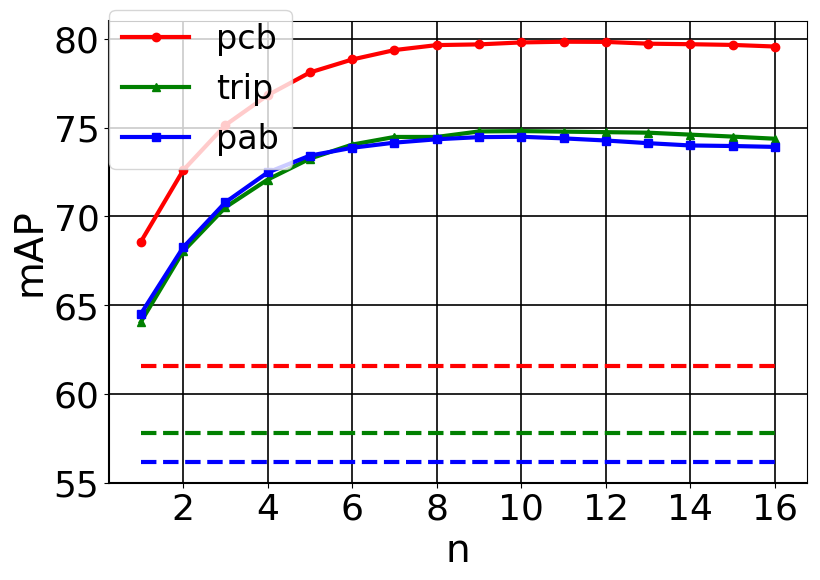}
    \includegraphics[width=1.62in]{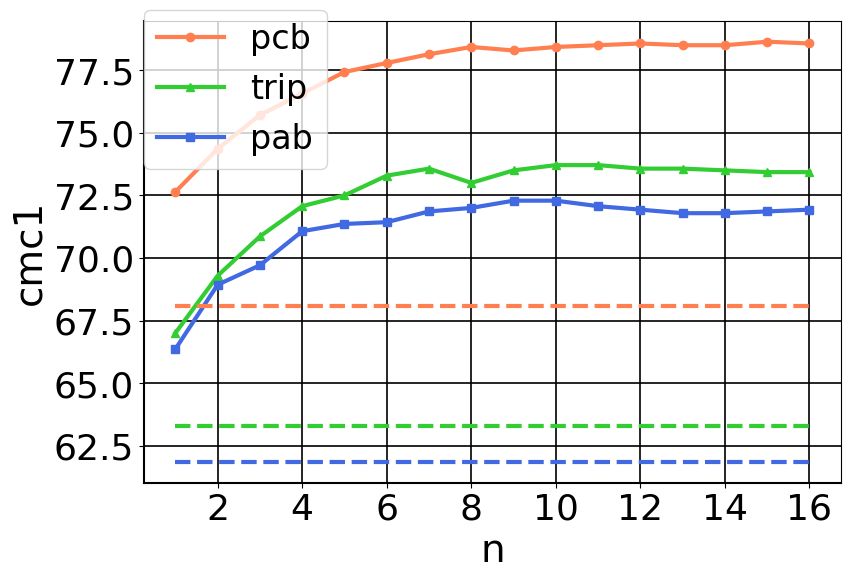}
    \includegraphics[width=1.62in]{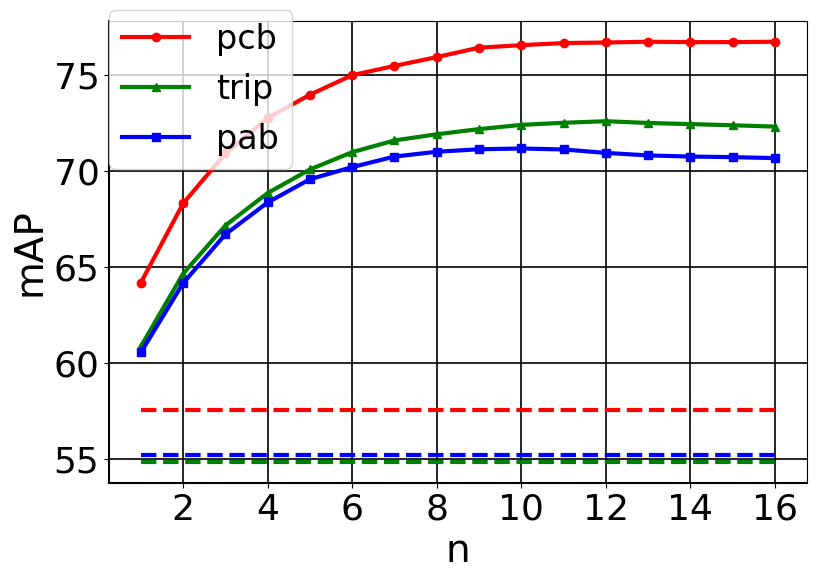}
    \includegraphics[width=1.62in]{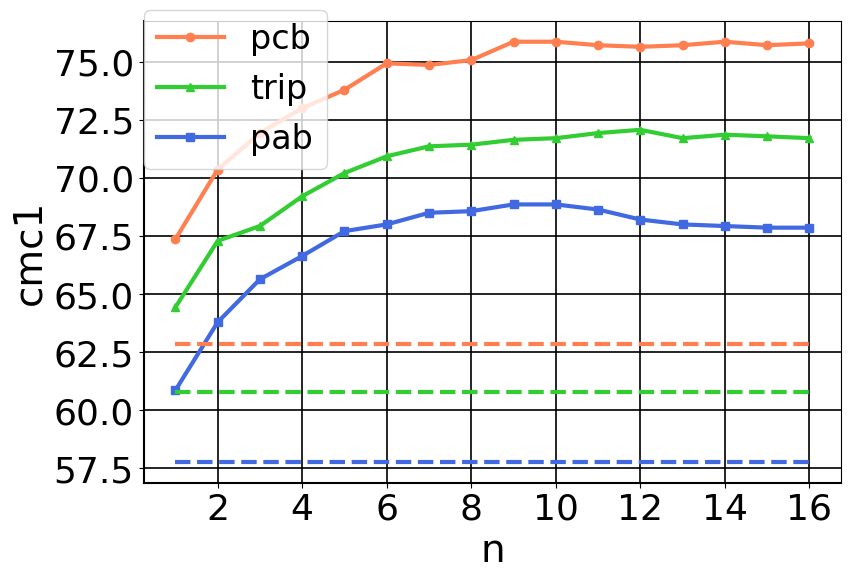}
    \caption{Performances (left: mAP; right cmc1) change with respect to $n$ on DukeMTMC, Market-1501, CUHK03-NP-lab, CUHK03-NP-det (from top to bottom) validation set, we fix hyper-parameters $\alpha$ and $K$ according to Table\ref{tab:duke}, \ref{tab:market} and \ref{tab:cuhk}. The dashed line is the baseline.}
    \label{fig:param_n}
\end{figure}

In Fig. \ref{fig:param_n}, we illustrate how the performance of mAP and cmc1 change with respect to $n$. Within a certain range, we can obtain better performance on both mAP and cmc1. However, more iterations do not always help to further improve performance. In fact, if we do not update gallery representations, the performance curves demonstrate saturating effects, meaning no new information is available under such a setting. However, since we are also updating the gallery representations, some noise information may be introduced to deteriorate the gallery impression with too many iterations. Specifically, for the CUHK03-NP (both labeled and detected) dataset, because the identities in gallery is same as those in query, its curves are more saturated for larger $n$. But for Market-1501 and DukeMTMC datasets, because of the existence of distractors in the gallery set, the performance will become worse for  large $n$. Besides, we find the optimal $n$ for cmc1 is about 2 iterations earlier than mAP, this may be because IIA can still have a considerable gain in recall but corrupt the top-1's identity within these iterations. Therefore, we recommend users to set $n$ in the range of $4 \sim 10$ for an unknown new dataset.

\subsubsection{Number of closest candidates $K$}
\begin{figure}[ht]
    \centering
    \includegraphics[width=1.62in]{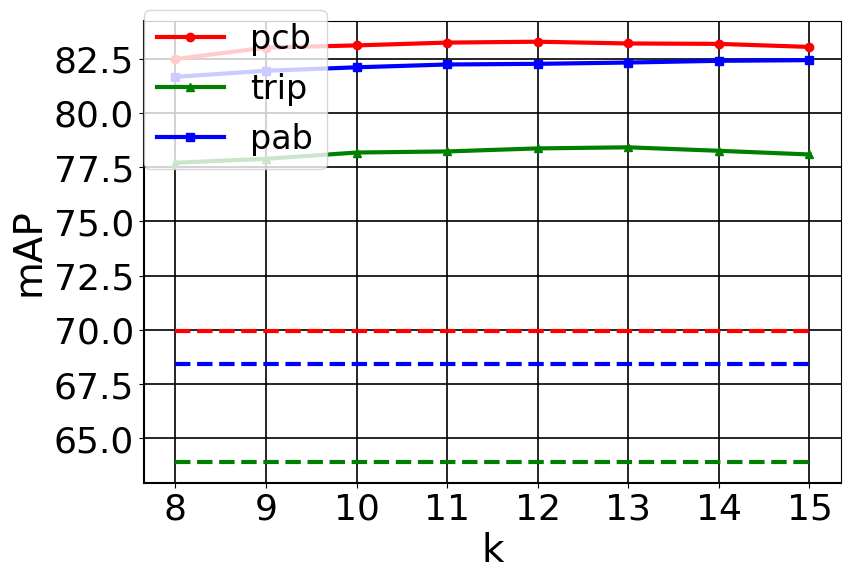}
    \includegraphics[width=1.62in]{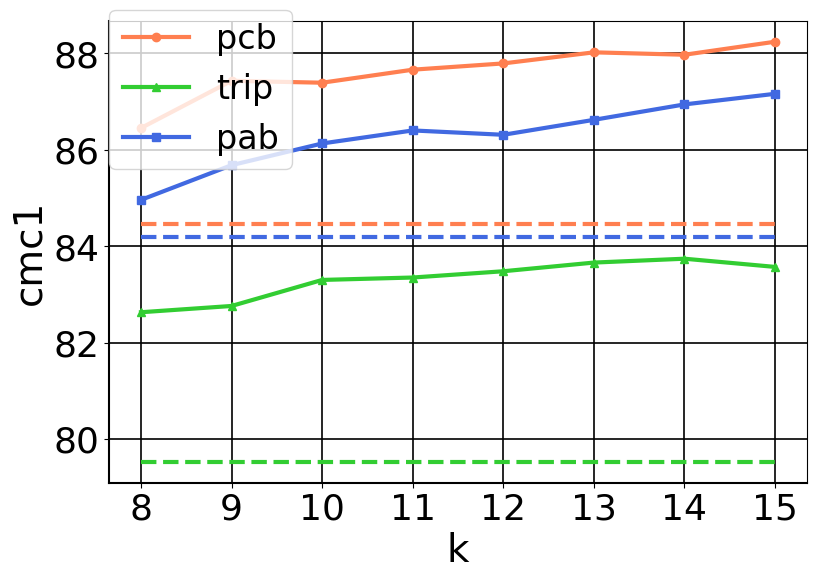}
    \includegraphics[width=1.62in]{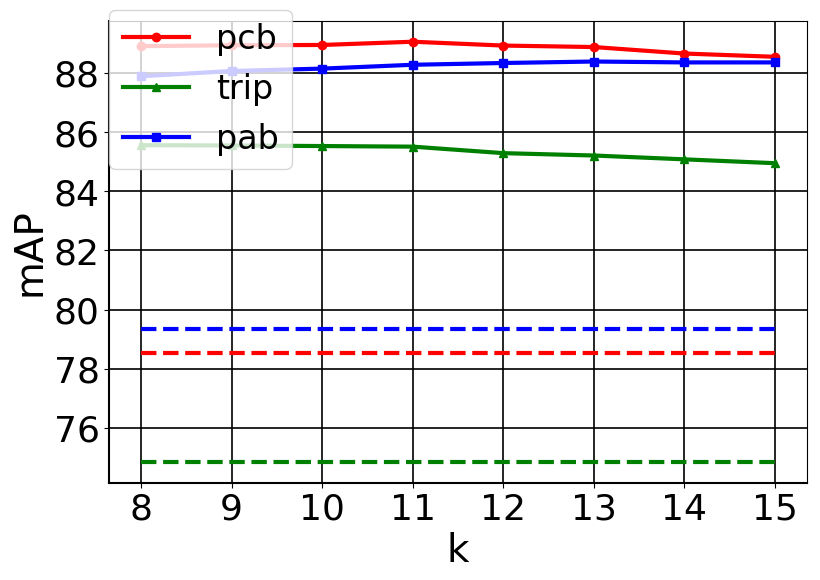}
    \includegraphics[width=1.62in]{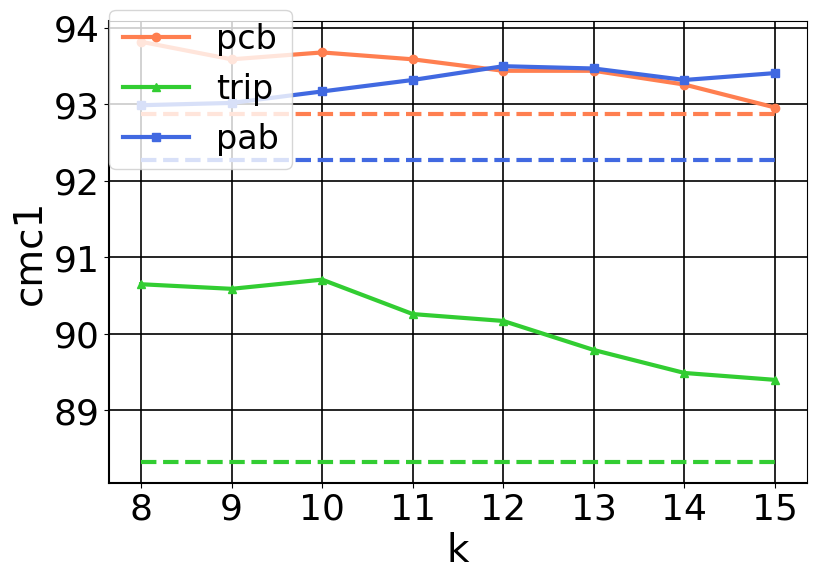}
    \includegraphics[width=1.62in]{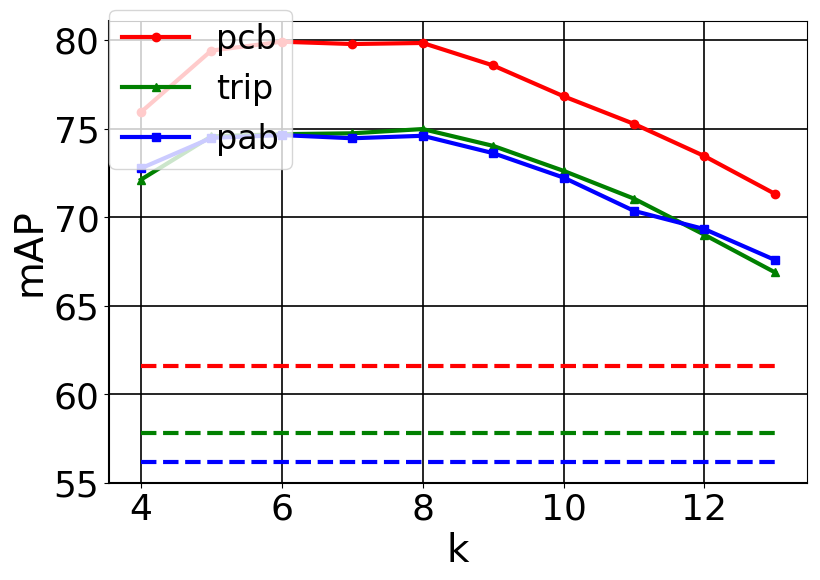}
    \includegraphics[width=1.62in]{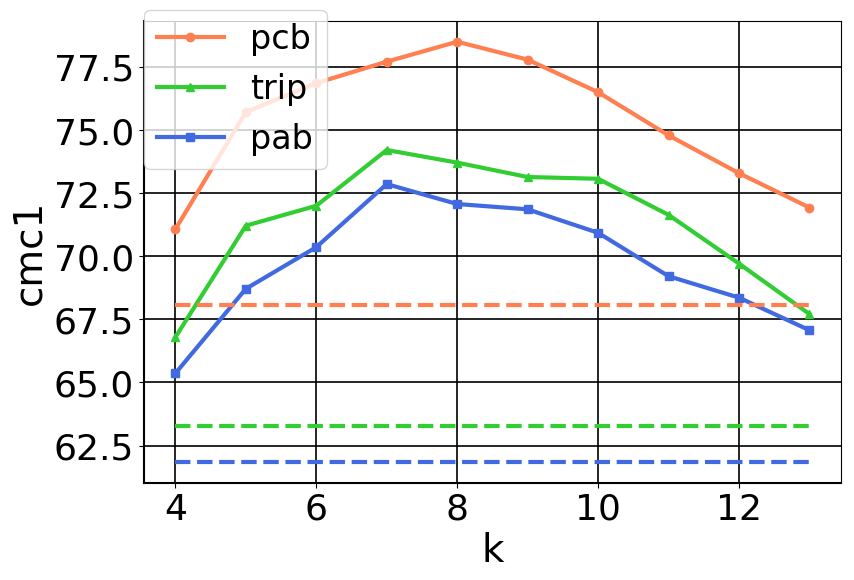}
    \includegraphics[width=1.62in]{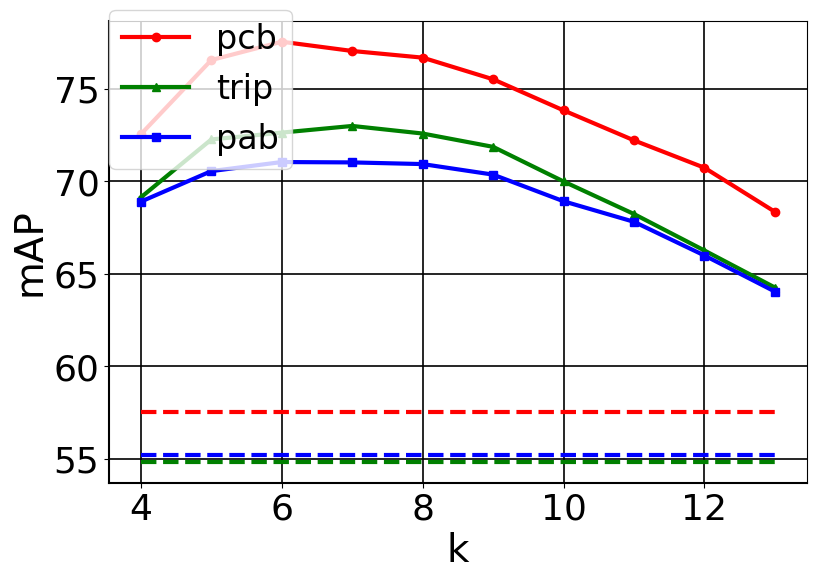}
    \includegraphics[width=1.62in]{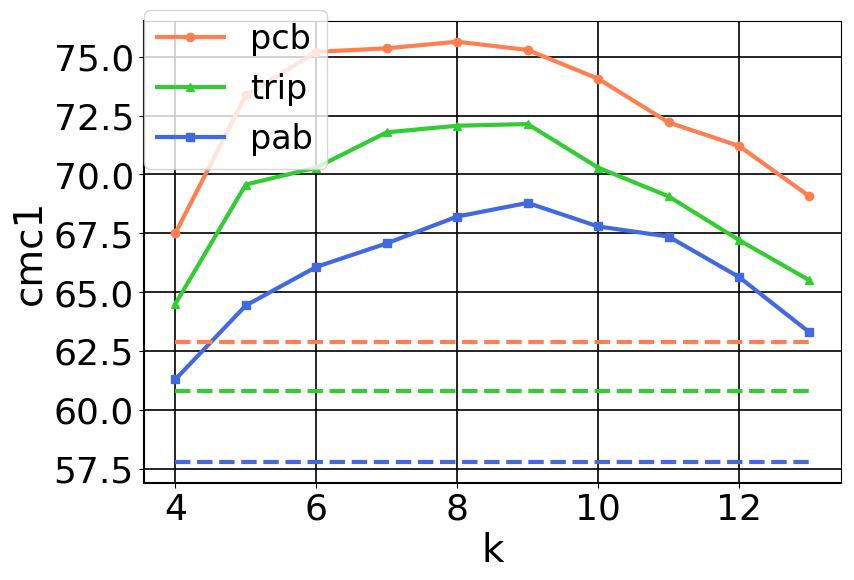}
    \caption{Performances (left: mAP; right cmc1) change with respect to $K$ on DukeMTMC, Market-1501, CUHK03-NP-lab, CUHK03-NP-det (from top to bottom) validation set, we fix hyper-parameters $\alpha$ and $n$ according to Table\ref{tab:duke}, \ref{tab:market} and \ref{tab:cuhk}. The dashed line is the baseline.}
    \label{fig:param_k}
\end{figure}

In our formulation, we only use the top-k nearest neighbors for impression updating. Here, we study the performance change with different neighbor number $K$ in Fig. \ref{fig:param_k}. It can be seen that, for the DukeMTMC and Market-1501 dataset, the mAP curves are very robust for a very wide range of $K$, while for CUHK03-NP the mAP drops quickly when $K$ surpasses a threshold ($K=8$), which means $K$ cannot be very large. In fact, from Table \ref{tab:dataset} we can find the number of person for per gallery identity is 15.9, 17.5 and 7.6 for DukeMTMC, Market-1501 and CUHK03-NP respectively. Intuitively, $K$ needs to be set as a moderate value. Too large $K$ means there must be some wrong candidates involved in query updating, while too small $K$ means much useful information is unexplored. We recommend $K$ to be the average number of samples per identity of a dataset or a little smaller than the average number, because not all the top-k ones are the right candidates.

\subsubsection{Temperature factor $\tau$}
\begin{figure}[ht]
    \centering
    \includegraphics[width=1.62in]{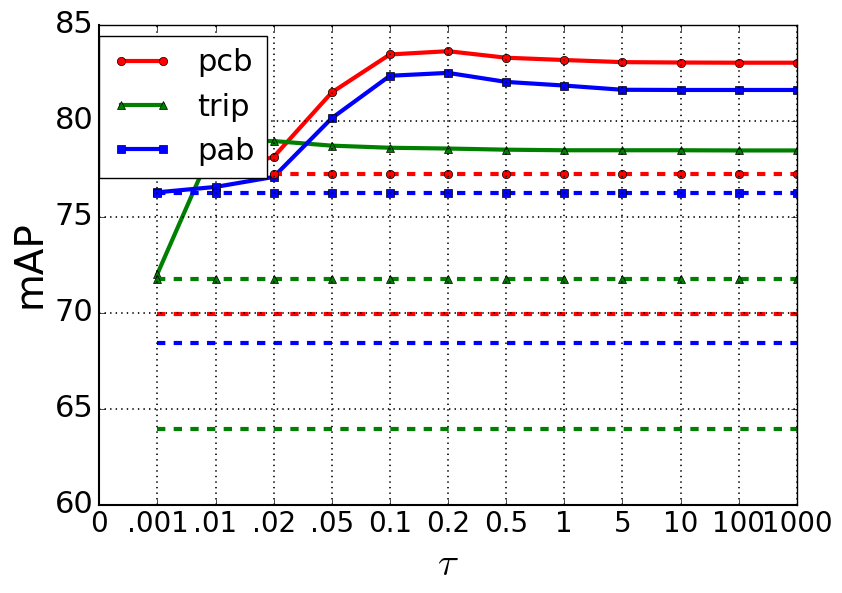}
    \includegraphics[width=1.62in]{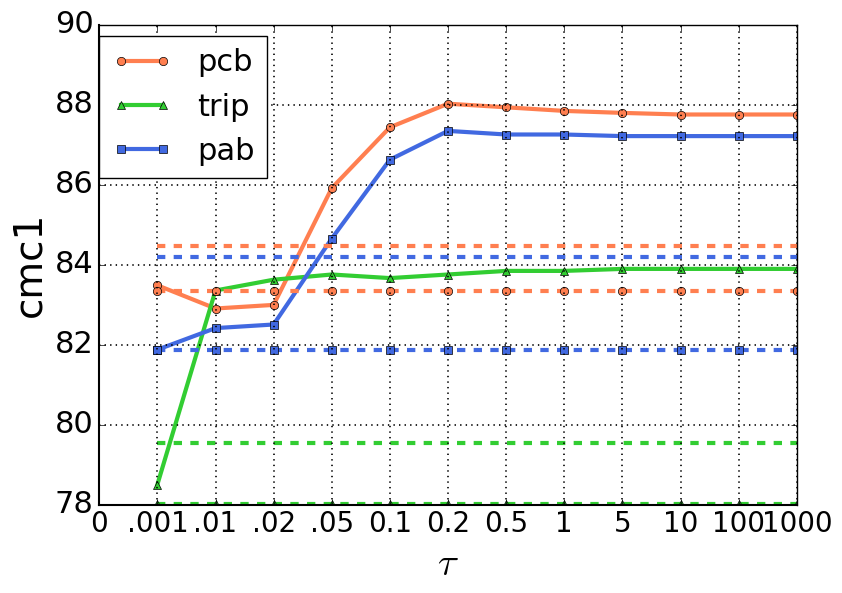}
    \includegraphics[width=1.62in]{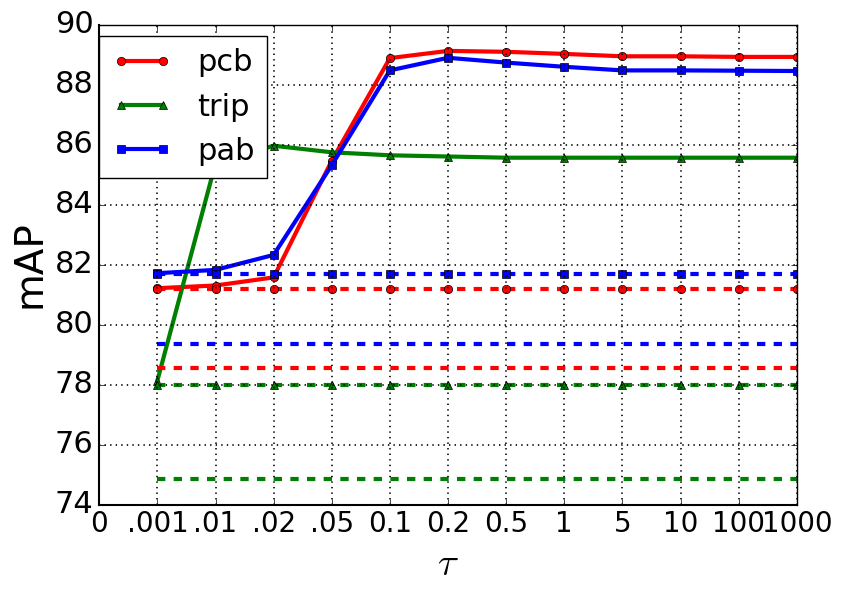}
    \includegraphics[width=1.62in]{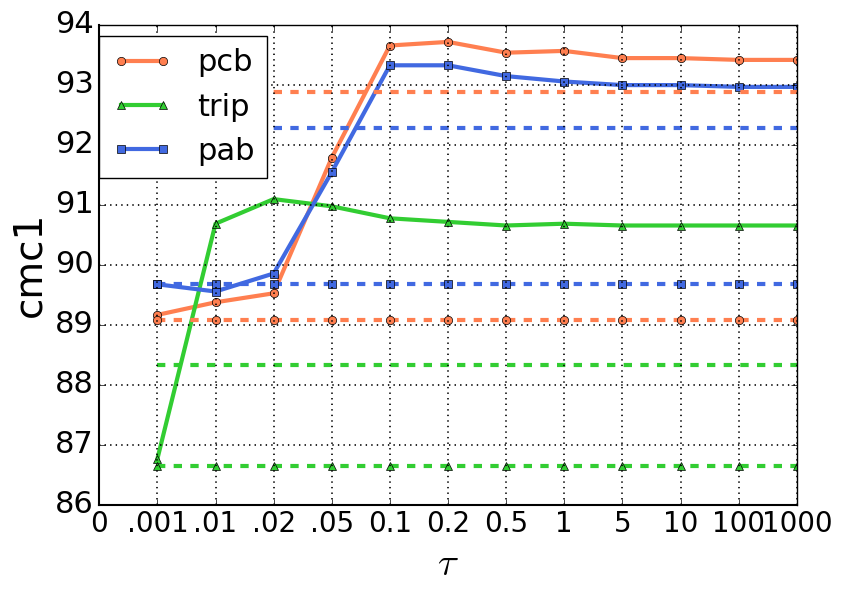}
    \includegraphics[width=1.62in]{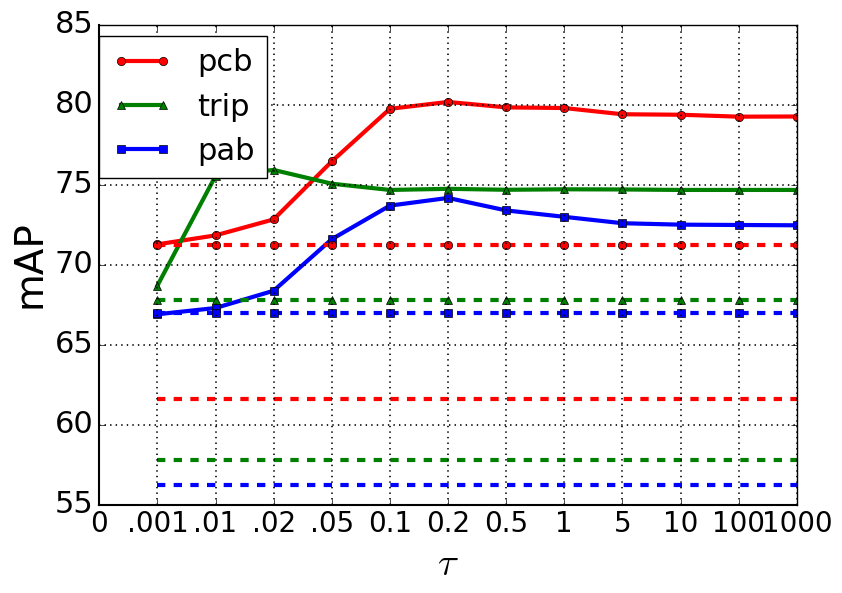}
    \includegraphics[width=1.62in]{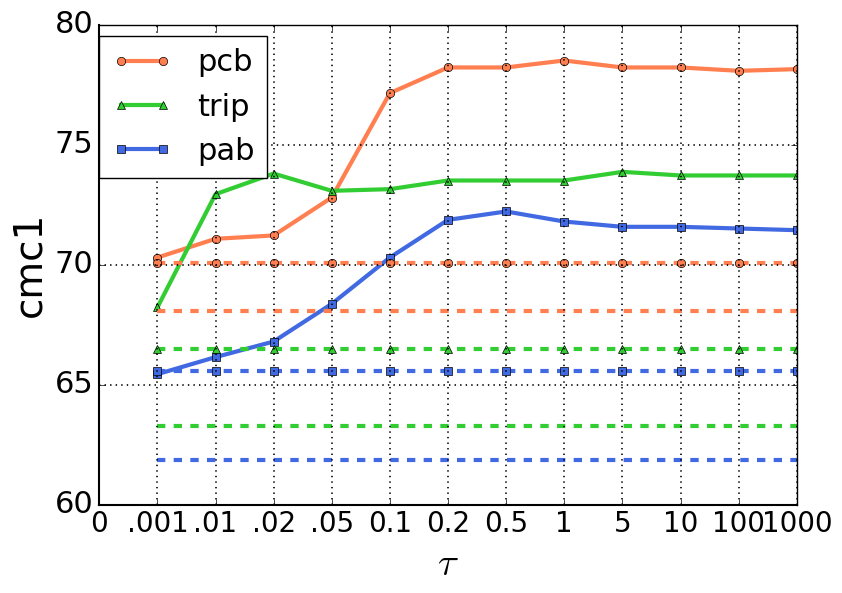}
    \includegraphics[width=1.62in]{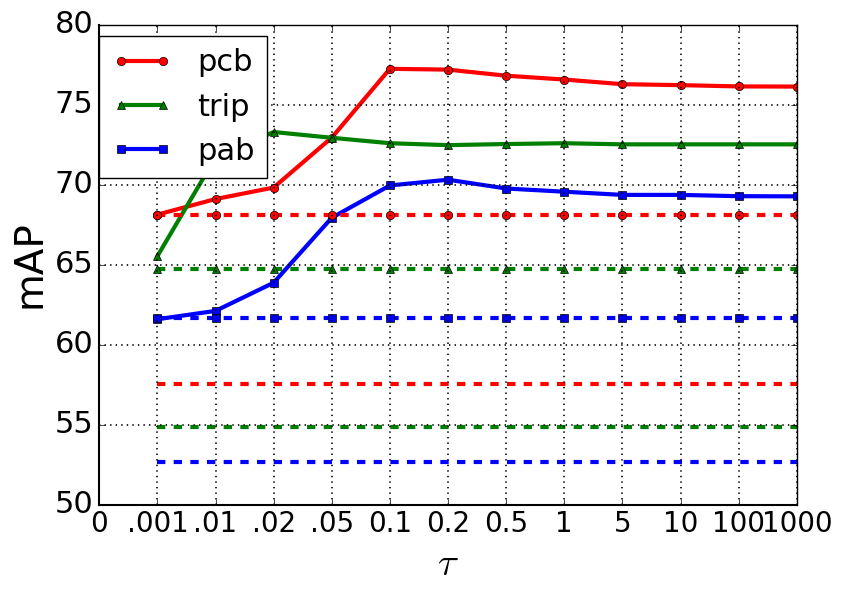}
    \includegraphics[width=1.62in]{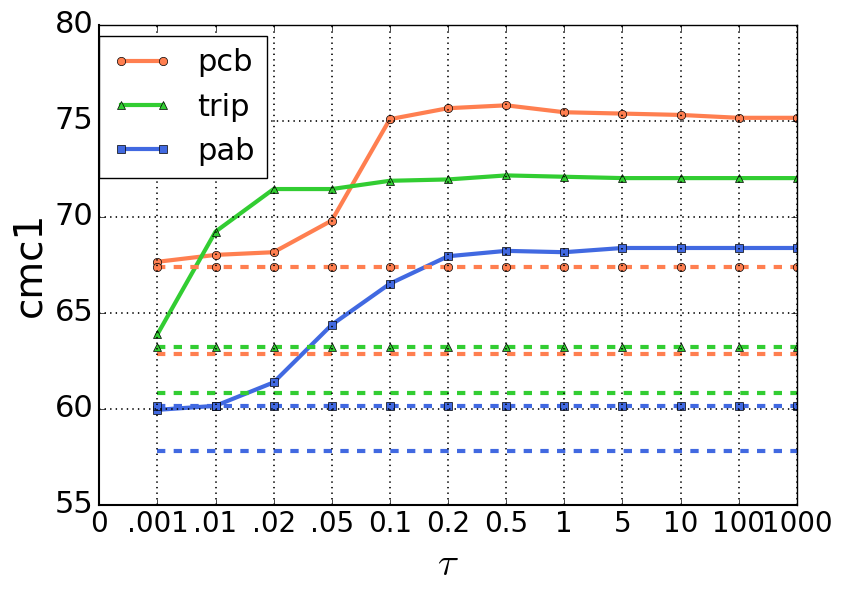}
    \caption{Performances (left: mAP; right cmc1) change with respect to $\tau$ on DukeMTMC, Market-1501, CUHK03-NP-lab, CUHK03-NP-det (from top to bottom) validation set, we fix hyper-parameters $\alpha$, $K$ and $n$ according to Table\ref{tab:duke}, \ref{tab:market} and \ref{tab:cuhk}. The dashed line is baseline, the dashed line with marker is for $K=1$.}
    \label{fig:param_tau}
\end{figure}

In Fig. \ref{fig:param_tau}, we illustrate how mAP and cmc1 change with respect to $\tau$. Overall, it shows that the performance is very stable with the change of $\tau$ especially when $\tau$ is larger than a threshold, and the optimal $\tau$ is consistently around $0.1\sim0.2$. Intuitively, when $\tau$ is small, the lower part of the top-k candidates' response will be shrunk to a very small value or even disappear. In the extreme case when $\tau$ is very small, only the top-1 candidate gallery will have a valid response. This indeed degenerates to the case that the number of closest candidates $K=1$, which determines the lower bound when $\tau$ changes. We also show the curve for $K=1$ (dashed line with marker) as a comparison and observe that it is sightly better than baseline but worse than the optimal $\tau$ with a large margin. In this paper, we set $\tau=0.2$ by default and recommend this setting to users. 

Note that, overall, most curves of the performance with respect to the hyper-parameters have a flipped-U shape, indicating an optimal position. We choose the optimal position according to the mAP metric since it considers both precision and recall. As shown in these figures and the above analysis, the performance of IIA is robust to hyper-parameters' tuning to some extent, demonstrating its practical applicability and effectiveness.

\subsection{Comparison to State-of-the-Arts}
\label{ssec:cts}

\begin{table*}
    \centering
    \caption{Comparison with state-of-the-art methods. We report the results of st-ReID obtained from the official released codes and models, which are lower than the paper reported ones. Res152 and Den201 mean the backbone networks are ResNet-152 and DenseNet-201, while the others are all based on ResNet-50. Missing entries mean that those paper didn't report that kind of results.}
    \begin{tabular}{ p{30mm} | >{\centering\arraybackslash}p{12.5mm} | >{\centering\arraybackslash}p{12.5mm} | >{\centering\arraybackslash}p{12.5mm} | >{\centering\arraybackslash}p{12.5mm} | >{\centering\arraybackslash}p{12.5mm} | >{\centering\arraybackslash}p{12.5mm} | >{\centering\arraybackslash}p{12.5mm} | >{\centering\arraybackslash}p{12.5mm} }
        \hline
        \multirow{2}{*}{Method} &
        \multicolumn{2}{ >{\centering\arraybackslash}p{25mm} | }{DukeMTMC-reID} & \multicolumn{2}{ >{\centering\arraybackslash}p{25mm} | }{Market-1501} &  \multicolumn{2}{ >{\centering\arraybackslash}p{25mm} | }{CUHK03-NP-lab} & \multicolumn{2}{ >{\centering\arraybackslash}p{25mm} }{CUHK03-NP-det} \\
        \cline{2-9} & mAP & cmc1 & mAP & cmc1 & mAP & cmc1 & mAP & cmc1 \\
        \hline
        
        \hline
        RW \cite{shen2018deep} & $66.4$ & $80.7$ & $82.5$ & $92.7$ & - & - & - & - \\
        GroupCRF \cite{chen2018group} & $69.5$ & $84.9$ & $81.6$ & $93.5$ & - & - & - & - \\
        MGN \cite{wang2018learning} & $78.4$ & $88.7$ & $86.9$ & $95.7$ & $66.0$ & $66.8$ & $67.4$ & $68.0$ \\
        MGN+RR \cite{wang2018learning} & - & - & $94.2$ & $96.6$ & - & - & - & - \\
        SPreID(Res152) \cite{kalayeh2018human} & $73.3$ & $85.9$ & $83.4$ & $93.7$ & - & - & - & -  \\
        SPreID(Res152)+RR \cite{kalayeh2018human} & $85.0$ & $89.0$ & $91.0$ & $94.6$ & - & - & - & -  \\
        DaRe(Den201) \cite{wang2018resource} & $64.5$ & $80.2$ & $76.0$ & $89.0$ & $61.6$ & $66.1$ & $59.0$ & $63.3$ \\
        DaRe(Den201)+RR \cite{wang2018resource} & $80.0$ & $84.4$ & $86.7$ & $90.9$ & $74.7$ & $73.8$ & $71.6$ & $70.6$ \\
        Mancs \cite{wang2018mancs} & $71.8$ & $84.9$ & $82.3$ & $93.1$ & $63.9$ & $69.0$ & $60.5$ & $65.5$ \\
        AACN \cite{xu2018attention} & $59.25$ & $76.84$ & $82.96$ & $88.69$ & $81.61$ & $81.86$ & $78.37$ & $79.14$ \\
        Pyramid \cite{zheng2019pyramidal} & $79.0$ & $89.0$ & $88.2$ & $95.7$ & $76.9$ & $78.9$ & $74.9$ & $78.9$ \\
        DSA \cite{zhang2019densely} & $74.3$ & $86.2$ & $87.6$ & $95.7$ & $75.2$ & $78.9$ & $73.1$ & $78.2$ \\
        SFT \cite{luo2019spectral} & $73.2$ & $86.9$ & $82.7$ & $93.4$ & $62.4$ & $68.2$ & - & - \\
        SFT+RR \cite{luo2019spectral} & $83.3$ & $88.3$ & $90.6$ & $93.5$ & $68.7$ & $71.4$ & - & - \\
        st-ReID{*} \cite{wang2019spatial} & $83.7$ & $94.3$ & $89.0$ & $98.1$ & - & - & - & - \\
        st-ReID+RR{*} \cite{wang2019spatial} & $91.4$ & $92.2$ & $95.0$ & $97.8$ & - & - & - & - \\
        Ours+MGN & $90.71$ & $92.24$ & $94.50$ & $95.69$ & $\mathbf{86.58}$ & $\mathbf{84.93}$ & $\mathbf{82.72}$ & $\mathbf{80.14}$ \\
        Ours+st-ReID & $\mathbf{91.8}$ & $\mathbf{94.4}$ & $\mathbf{96.0}$ & $\mathbf{98.2}$ & - & - & - & - \\
        \hline
    \end{tabular}
    \label{tab:ss}
\end{table*}

In this section, we collect our best performing model based on two baseline models and compare our results with many latest state-of-the-art models including: RW \cite{shen2018deep}, GroupCRF \cite{chen2018group}, MGN \cite{wang2018learning}, SPreID \cite{kalayeh2018human}, 
DaRe \cite{wang2018resource}, Mancs \cite{wang2018mancs}, AACN \cite{xu2018attention}, DSA \cite{zhang2019densely}, Pyramid \cite{zheng2019pyramidal}, SFT \cite{luo2019spectral} and st-ReID \cite{wang2019spatial}. As shown in Table \ref{tab:ss}, when combining the proposed IIA with MGN\footnote{We train MGN with our re-implementation, whose performance is slightly worse than the official reported one, more details can be found in Table\ref{tab:duke},\ref{tab:market},\ref{tab:cuhk}.} or st-ReID\footnote{We evaluate with st-ReID official released model, whose performance is slightly different from paper reported, please refer to \url{https://github.com/Wanggcong/Spatial-Temporal-Re-identification} for details.}, we can achieve better results than all these state-of-the-art models. Especially on the CUHK03 dataset, our method surpasses existing methods by a large margin.

\subsection{Limitation and Failure cases}
\label{ssec:limitation_and_failurecases} 
\begin{figure}[ht]
    \centering
    \includegraphics[width=3.2in]{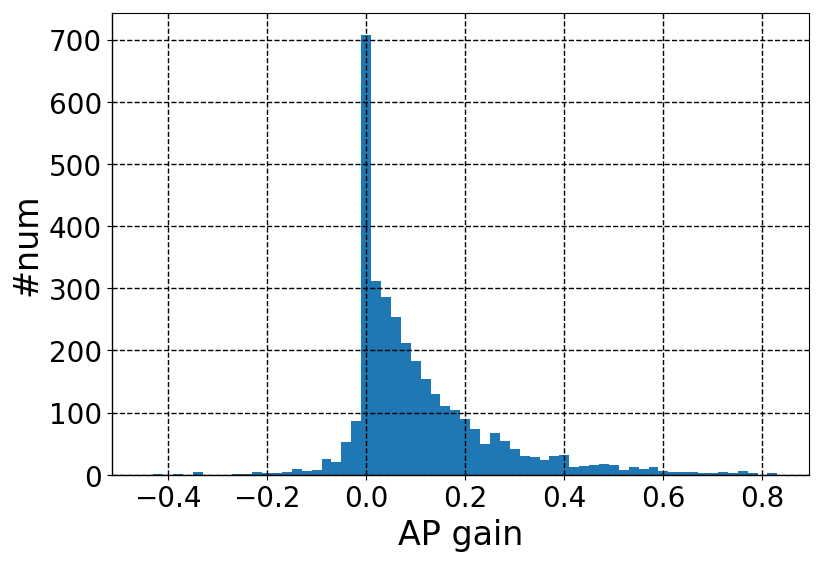}
    \caption{Histogram of AP gain for Market-1501 with PCB features after applying IIA.}
    \label{fig:hist_map}
\end{figure}

\begin{figure}
    \centering
    \subfigure[Failure case 1, initial ranking list is a mess.] {
		\includegraphics[width=3.2in]{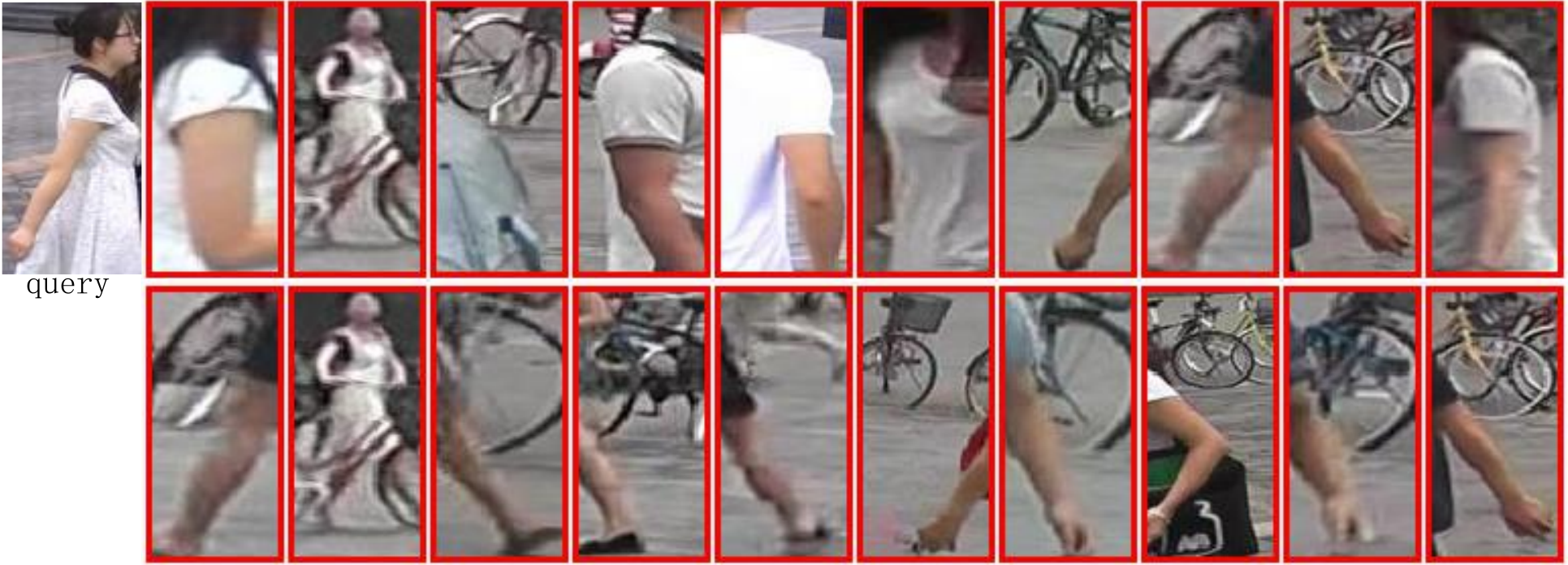}
		\label{fig:f1}
	}
	\subfigure[Failure cases 2, most of the false candidates belong to another person.] {
		\includegraphics[width=3.2in]{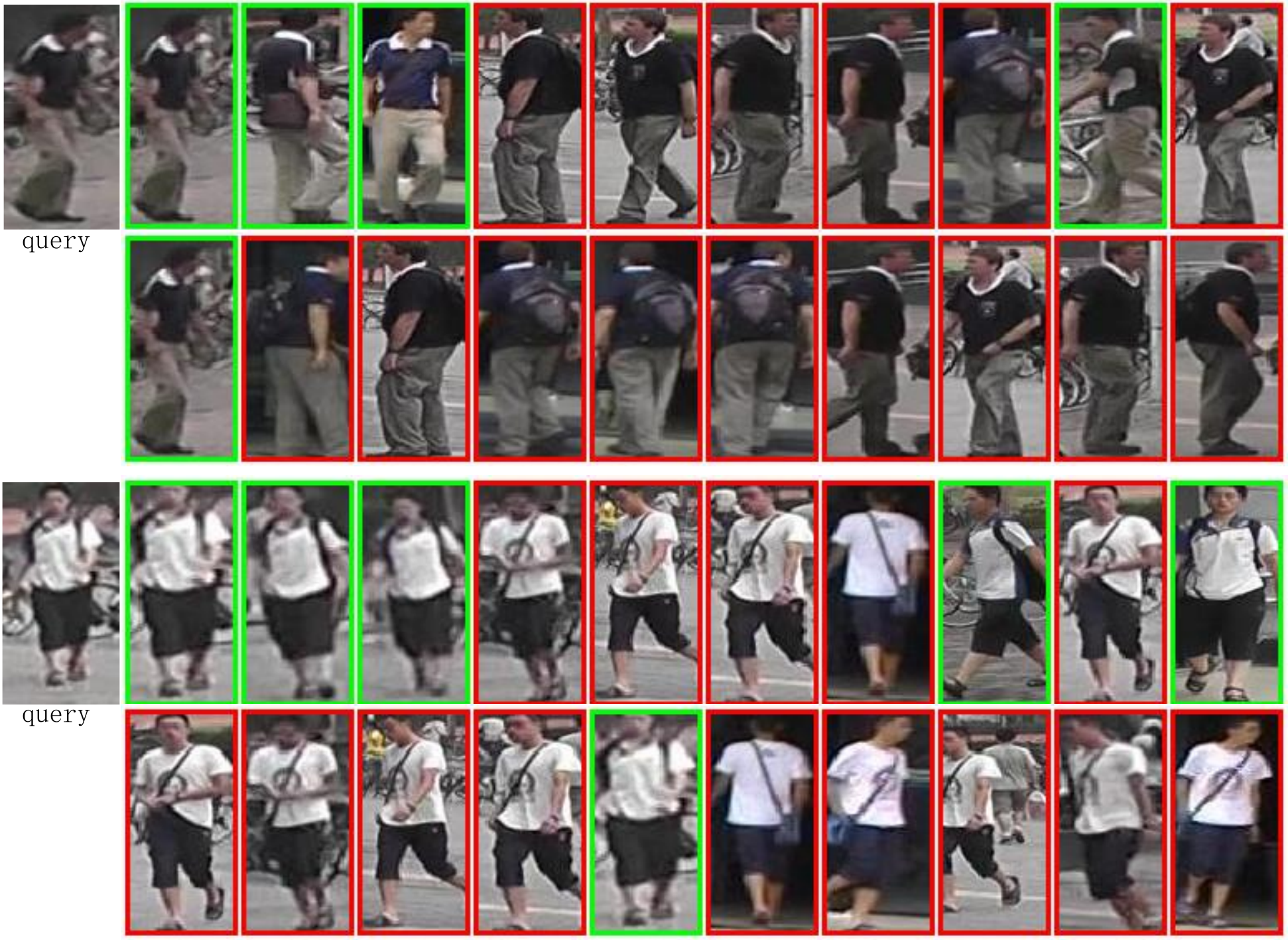}
		\label{fig:f2}
	}
	\subfigure[Failure case 3, the first top few candidates are not correct.] {
		\includegraphics[width=3.2in]{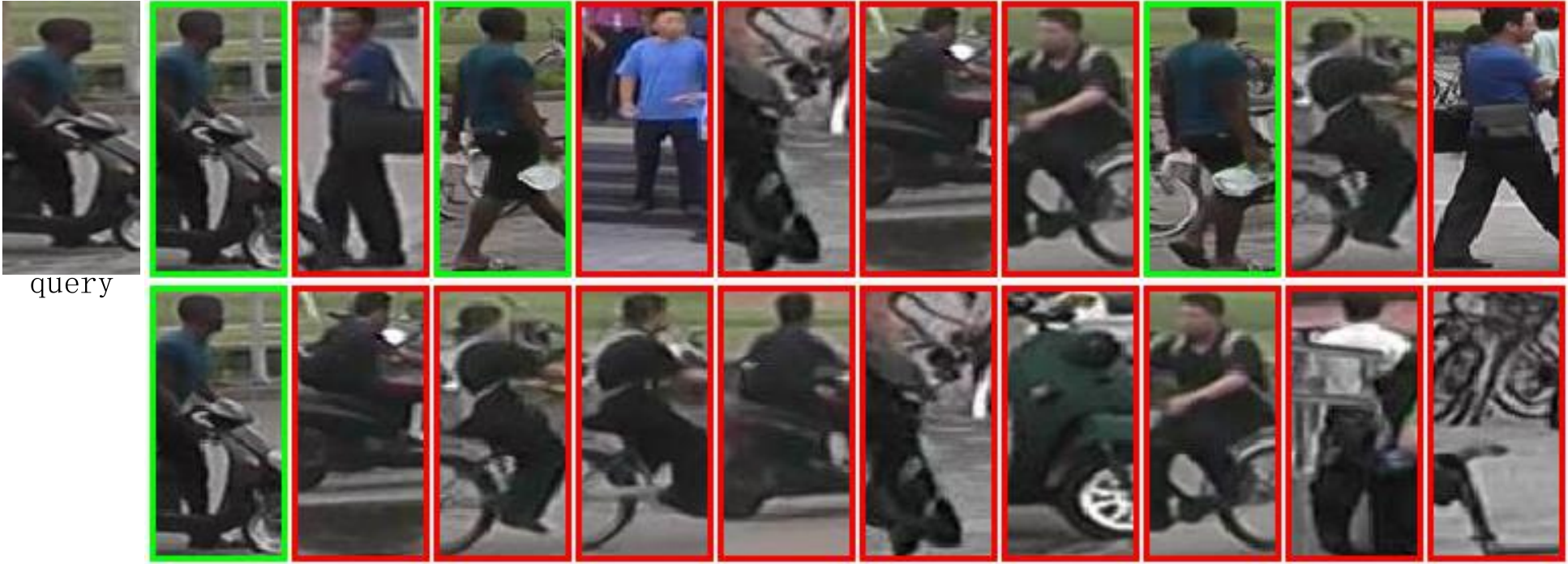}
		\label{fig:f3}
	}
    \caption{Example of failure cases. For each query image, the first row shows the original ranking list, while the second row demonstrates the updated ranking results obtained by IIA$_{bas}$. Images surrounded by green box denotes its identity is the same with that of the query image and the red-boxed images are negative ones.}
    \label{fig:failure1}
\end{figure}

Despite the superiority of IIA, it also has some failure cases. Fig. \ref{fig:hist_map} shows the gain of AP after applying IIA for Market-1501 with PCB \cite{sun2018PCB} features, from which we can see for most of the queries, their APs will be enhanced, but there are still a small portion where the value of APs decrease. After exploring these failure cases, we found the following  three typical cases when IIA performs relatively poorly.
\begin{itemize}
    \item The initial ranking list is very messy, e.g. none of the candidates                             is correct. Fig.\ref{fig:f1} shows this example.
    \item There are a lot of false candidates belonging to the same other person in $K$ closest candidates, and this is the most common failure case. We show such examples in  Fig.\ref{fig:f2}.
    \item The first top few candidates in the initial rank are not the same as the query. Fig. \ref{fig:f3} show this example
\end{itemize}

The latter two failure cases share the same essence: Error Propagation. This is due to the inherent design philosophy of IIA that tries to pass information from gallery to query.  Potential solutions include but not limited to, e.g., making an additional design to decide whether to update. For example, if we find the top-k candidates' feature can be divided into 2 clusters almost uniformly, we directly discard this update.

\section{Conclusion}
\label{sec:concl}

In this paper, we shed light on a natural iterative impression update process. Intuitively, we put our impression about one person updates after we see more aspects of this person computationally into practice. We incorporate this idea to person re-identification with an iterative attentional aggregation model, where the representation of a query image is iteratively updated with new information from those closely related candidates in the gallery. The proposed model is flexible to leverage different representations and similarity metrics, and therefore results in both a simple practical solution to largely improve the performance of existing re-ID models and also a strong re-ID model that achieves state-of-the-art performance on standard benchmarks including CUHK03, Market-1501 and DukeMTMC. The model also naturally supports certain ensemble designs using representations at each iteration and potential fine-tuning to original features in an end-to-end manner, both of which could potentially further strengthen the model and will be our focus in the near future.

\bibliographystyle{IEEEtran}
\bibliography{egbib}

\begin{thebibliography}{10}
\providecommand{\url}[1]{#1}
\csname url@samestyle\endcsname
\providecommand{\newblock}{\relax}
\providecommand{\bibinfo}[2]{#2}
\providecommand{\BIBentrySTDinterwordspacing}{\spaceskip=0pt\relax}
\providecommand{\BIBentryALTinterwordstretchfactor}{4}
\providecommand{\BIBentryALTinterwordspacing}{\spaceskip=\fontdimen2\font plus
\BIBentryALTinterwordstretchfactor\fontdimen3\font minus
  \fontdimen4\font\relax}
\providecommand{\BIBforeignlanguage}[2]{{%
\expandafter\ifx\csname l@#1\endcsname\relax
\typeout{** WARNING: IEEEtran.bst: No hyphenation pattern has been}%
\typeout{** loaded for the language `#1'. Using the pattern for}%
\typeout{** the default language instead.}%
\else
\language=\csname l@#1\endcsname
\fi
#2}}
\providecommand{\BIBdecl}{\relax}
\BIBdecl

\bibitem{wang2007shape}
X.~Wang, G.~Doretto, T.~Sebastian, J.~Rittscher, and P.~Tu, ``Shape and
  appearance context modeling,'' 2007.

\bibitem{liu2017end}
H.~Liu, J.~Feng, M.~Qi, J.~Jiang, and S.~Yan, ``End-to-end comparative
  attention networks for person re-identification,'' \emph{IEEE Transactions on
  Image Processing}, vol.~26, no.~7, pp. 3492--3506, 2017.

\bibitem{sun2018PCB}
Y.~Sun, L.~Zheng, Y.~Yang, Q.~Tian, and S.~Wang, ``Beyond part models: Person
  retrieval with refined part pooling (and a strong convolutional baseline),''
  in \emph{ECCV}, 2018.

\bibitem{chen2018group}
D.~Chen, D.~Xu, H.~Li, N.~Sebe, and X.~Wang, ``Group consistent similarity
  learning via deep crf for person re-identification,'' in \emph{Proceedings of
  the IEEE Conference on Computer Vision and Pattern Recognition}, 2018, pp.
  8649--8658.

\bibitem{shen2018deep}
Y.~Shen, H.~Li, T.~Xiao, S.~Yi, D.~Chen, and X.~Wang, ``Deep group-shuffling
  random walk for person re-identification,'' in \emph{Proceedings of the IEEE
  Conference on Computer Vision and Pattern Recognition}, 2018, pp. 2265--2274.

\bibitem{suh2018part}
Y.~Suh, J.~Wang, S.~Tang, T.~Mei, and K.~M. Lee, ``Part-aligned bilinear
  representations for person re-identification,'' \emph{arXiv preprint
  arXiv:1804.07094}, 2018.

\bibitem{zheng2019joint}
Z.~Zheng, X.~Yang, Z.~Yu, L.~Zheng, Y.~Yang, and J.~Kautz, ``Joint
  discriminative and generative learning for person re-identification,'' in
  \emph{Proceedings of the IEEE Conference on Computer Vision and Pattern
  Recognition}, 2019, pp. 2138--2147.

\bibitem{eom2019learning}
C.~Eom and B.~Ham, ``Learning disentangled representation for robust person
  re-identification,'' in \emph{Advances in Neural Information Processing
  Systems}, 2019, pp. 5298--5309.

\bibitem{koestinger2012large}
M.~Koestinger, M.~Hirzer, P.~Wohlhart, P.~M. Roth, and H.~Bischof, ``Large
  scale metric learning from equivalence constraints,'' in \emph{Computer
  Vision and Pattern Recognition (CVPR), 2012 IEEE Conference on}.\hskip 1em
  plus 0.5em minus 0.4em\relax IEEE, 2012, pp. 2288--2295.

\bibitem{weinberger2009distance}
K.~Q. Weinberger and L.~K. Saul, ``Distance metric learning for large margin
  nearest neighbor classification,'' \emph{Journal of Machine Learning
  Research}, vol.~10, no. Feb, pp. 207--244, 2009.

\bibitem{xiong2014person}
F.~Xiong, M.~Gou, O.~Camps, and M.~Sznaier, ``Person re-identification using
  kernel-based metric learning methods,'' in \emph{European conference on
  computer vision}.\hskip 1em plus 0.5em minus 0.4em\relax Springer, 2014, pp.
  1--16.

\bibitem{liao2015person}
S.~Liao, Y.~Hu, X.~Zhu, and S.~Z. Li, ``Person re-identification by local
  maximal occurrence representation and metric learning,'' in \emph{Proceedings
  of the IEEE conference on computer vision and pattern recognition}, 2015, pp.
  2197--2206.

\bibitem{martinel2015kernelized}
N.~Martinel, C.~Micheloni, and G.~L. Foresti, ``Kernelized saliency-based
  person re-identification through multiple metric learning,'' \emph{IEEE
  Transactions on Image Processing}, vol.~24, no.~12, pp. 5645--5658, 2015.

\bibitem{paisitkriangkrai2015learning}
S.~Paisitkriangkrai, C.~Shen, and A.~Van Den~Hengel, ``Learning to rank in
  person re-identification with metric ensembles,'' in \emph{Proceedings of the
  IEEE Conference on Computer Vision and Pattern Recognition}, 2015, pp.
  1846--1855.

\bibitem{oh2016deep}
H.~Oh~Song, Y.~Xiang, S.~Jegelka, and S.~Savarese, ``Deep metric learning via
  lifted structured feature embedding,'' in \emph{Proceedings of the IEEE
  Conference on Computer Vision and Pattern Recognition}, 2016, pp. 4004--4012.

\bibitem{yang2017person}
X.~Yang, M.~Wang, and D.~Tao, ``Person re-identification with metric learning
  using privileged information,'' \emph{IEEE Transactions on Image Processing},
  vol.~27, no.~2, pp. 791--805, 2017.

\bibitem{wang2018learning}
G.~Wang, Y.~Yuan, X.~Chen, J.~Li, and X.~Zhou, ``Learning discriminative
  features with multiple granularities for person re-identification,'' in
  \emph{2018 ACM Multimedia Conference on Multimedia Conference}.\hskip 1em
  plus 0.5em minus 0.4em\relax ACM, 2018, pp. 274--282.

\bibitem{zhang2020ordered}
L.~Zhang, Z.~Shi, J.~T. Zhou, M.-M. Cheng, Y.~Liu, J.-W. Bian, Z.~Zeng, and
  C.~Shen, ``Ordered or orderless: A revisit for video based person
  re-identification,'' \emph{IEEE Transactions on Pattern Analysis and Machine
  Intelligence}, 2020.

\bibitem{jegou2007contextual}
H.~Jegou, H.~Harzallah, and C.~Schmid, ``A contextual dissimilarity measure for
  accurate and efficient image search,'' in \emph{Computer Vision and Pattern
  Recognition, 2007. CVPR'07. IEEE Conference on}.\hskip 1em plus 0.5em minus
  0.4em\relax IEEE, 2007, pp. 1--8.

\bibitem{qin2011hello}
D.~Qin, S.~Gammeter, L.~Bossard, T.~Quack, and L.~Van~Gool, ``Hello neighbor:
  Accurate object retrieval with k-reciprocal nearest neighbors,'' in
  \emph{Computer Vision and Pattern Recognition (CVPR), 2011 IEEE Conference
  on}.\hskip 1em plus 0.5em minus 0.4em\relax IEEE, 2011, pp. 777--784.

\bibitem{ma2015cross}
A.~J. Ma, J.~Li, P.~C. Yuen, and P.~Li, ``Cross-domain person reidentification
  using domain adaptation ranking svms,'' \emph{IEEE transactions on image
  processing}, vol.~24, no.~5, pp. 1599--1613, 2015.

\bibitem{bai2016sparse}
S.~Bai and X.~Bai, ``Sparse contextual activation for efficient visual
  re-ranking,'' \emph{IEEE Transactions on Image Processing}, vol.~25, no.~3,
  pp. 1056--1069, 2016.

\bibitem{garcia2017discriminant}
J.~Garcia, N.~Martinel, A.~Gardel, I.~Bravo, G.~L. Foresti, and C.~Micheloni,
  ``Discriminant context information analysis for post-ranking person
  re-identification,'' \emph{IEEE Transactions on Image Processing}, vol.~26,
  no.~4, pp. 1650--1665, 2017.

\bibitem{zhong2017re}
Z.~Zhong, L.~Zheng, D.~Cao, and S.~Li, ``Re-ranking person re-identification
  with k-reciprocal encoding,'' in \emph{Computer Vision and Pattern
  Recognition (CVPR), 2017 IEEE Conference on}.\hskip 1em plus 0.5em minus
  0.4em\relax IEEE, 2017, pp. 3652--3661.

\bibitem{mende2013neural}
P.~Mende-Siedlecki, Y.~Cai, and A.~Todorov, ``The neural dynamics of updating
  person impressions,'' \emph{Social cognitive and affective neuroscience},
  vol.~8, no.~6, pp. 623--631, 2013.

\bibitem{salton1990improving}
G.~Salton and C.~Buckley, ``Improving retrieval performance by relevance
  feedback,'' \emph{Journal of the American society for information science},
  vol.~41, no.~4, pp. 288--297, 1990.

\bibitem{zheng2016person}
L.~Zheng, Y.~Yang, and A.~G. Hauptmann, ``Person re-identification: Past,
  present and future,'' \emph{arXiv preprint arXiv:1610.02984}, 2016.

\bibitem{hermans2017defense}
A.~Hermans, L.~Beyer, and B.~Leibe, ``In defense of the triplet loss for person
  re-identification,'' \emph{arXiv preprint arXiv:1703.07737}, 2017.

\bibitem{shen2018person}
Y.~Shen, H.~Li, S.~Yi, D.~Chen, and X.~Wang, ``Person re-identification with
  deep similarity-guided graph neural network,'' in \emph{Proceedings of the
  European conference on computer vision (ECCV)}, 2018, pp. 486--504.

\bibitem{ye2016person}
M.~Ye, C.~Liang, Y.~Yu, Z.~Wang, Q.~Leng, C.~Xiao, J.~Chen, and R.~Hu, ``Person
  reidentification via ranking aggregation of similarity pulling and
  dissimilarity pushing,'' \emph{IEEE Transactions on Multimedia}, vol.~18,
  no.~12, pp. 2553--2566, 2016.

\bibitem{leng2015person}
Q.~Leng, R.~Hu, C.~Liang, Y.~Wang, and J.~Chen, ``Person re-identification with
  content and context re-ranking,'' \emph{Multimedia Tools and Applications},
  vol.~74, no.~17, pp. 6989--7014, 2015.

\bibitem{garcia2015person}
J.~Garcia, N.~Martinel, C.~Micheloni, and A.~Gardel, ``Person re-identification
  ranking optimisation by discriminant context information analysis,'' in
  \emph{Proceedings of the IEEE International Conference on Computer Vision},
  2015.

\bibitem{ye2015coupled}
M.~Ye, J.~Chen, Q.~Leng, C.~Liang, Z.~Wang, and K.~Sun, ``Coupled-view based
  ranking optimization for person re-identification,'' in \emph{International
  Conference on Multimedia Modeling}.\hskip 1em plus 0.5em minus 0.4em\relax
  Springer, 2015, pp. 105--117.

\bibitem{bai2019re}
S.~Bai, P.~Tang, P.~H. Torr, and L.~J. Latecki, ``Re-ranking via metric fusion
  for object retrieval and person re-identification,'' in \emph{Proceedings of
  the IEEE Conference on Computer Vision and Pattern Recognition}, 2019, pp.
  740--749.

\bibitem{chum2007total}
O.~Chum, J.~Philbin, J.~Sivic, M.~Isard, and A.~Zisserman, ``Total recall:
  Automatic query expansion with a generative feature model for object
  retrieval,'' in \emph{2007 IEEE 11th International Conference on Computer
  Vision}.\hskip 1em plus 0.5em minus 0.4em\relax IEEE, 2007, pp. 1--8.

\bibitem{arandjelovic2012three}
R.~Arandjelovi{\'c} and A.~Zisserman, ``Three things everyone should know to
  improve object retrieval,'' in \emph{2012 IEEE Conference on Computer Vision
  and Pattern Recognition}.\hskip 1em plus 0.5em minus 0.4em\relax IEEE, 2012,
  pp. 2911--2918.

\bibitem{green1984iteratively}
P.~J. Green, ``Iteratively reweighted least squares for maximum likelihood
  estimation, and some robust and resistant alternatives,'' \emph{Journal of
  the Royal Statistical Society: Series B (Methodological)}, vol.~46, no.~2,
  pp. 149--170, 1984.

\bibitem{xin2016maximal}
B.~Xin, Y.~Wang, W.~Gao, D.~Wipf, and B.~Wang, ``Maximal sparsity with deep
  networks?'' in \emph{Advances in Neural Information Processing Systems},
  2016, pp. 4340--4348.

\bibitem{daubechies2010iteratively}
I.~Daubechies, R.~DeVore, M.~Fornasier, and C.~S. G{\"u}nt{\"u}rk,
  ``Iteratively reweighted least squares minimization for sparse recovery,''
  \emph{Communications on Pure and Applied Mathematics: A Journal Issued by the
  Courant Institute of Mathematical Sciences}, vol.~63, no.~1, pp. 1--38, 2010.

\bibitem{vaswani2017attention}
A.~Vaswani, N.~Shazeer, N.~Parmar, J.~Uszkoreit, L.~Jones, A.~N. Gomez,
  {\L}.~Kaiser, and I.~Polosukhin, ``Attention is all you need,'' in
  \emph{Advances in Neural Information Processing Systems}, 2017, pp.
  5998--6008.

\bibitem{devlin2018bert}
J.~Devlin, M.-W. Chang, K.~Lee, and K.~Toutanova, ``Bert: Pre-training of deep
  bidirectional transformers for language understanding,'' \emph{arXiv preprint
  arXiv:1810.04805}, 2018.

\bibitem{weston2014memory}
J.~Weston, S.~Chopra, and A.~Bordes, ``Memory networks,'' \emph{arXiv preprint
  arXiv:1410.3916}, 2014.

\bibitem{sukhbaatar2015end}
S.~Sukhbaatar, J.~Weston, R.~Fergus \emph{et~al.}, ``End-to-end memory
  networks,'' in \emph{Advances in neural information processing systems},
  2015, pp. 2440--2448.

\bibitem{graves2014neural}
A.~Graves, G.~Wayne, and I.~Danihelka, ``Neural turing machines,'' \emph{arXiv
  preprint arXiv:1410.5401}, 2014.

\bibitem{zheng2017unlabeled}
Z.~Zheng, L.~Zheng, and Y.~Yang, ``Unlabeled samples generated by gan improve
  the person re-identification baseline in vitro,'' in \emph{Proceedings of the
  IEEE International Conference on Computer Vision}, 2017.

\bibitem{zheng2015scalable}
L.~Zheng, L.~Shen, L.~Tian, S.~Wang, J.~Wang, and Q.~Tian, ``Scalable person
  re-identification: A benchmark,'' in \emph{Computer Vision, IEEE
  International Conference on}, 2015.

\bibitem{felzenszwalb2008discriminatively}
P.~Felzenszwalb, D.~McAllester, and D.~Ramanan, ``A discriminatively trained,
  multiscale, deformable part model,'' 2008.

\bibitem{li2014deepreid}
W.~Li, R.~Zhao, T.~Xiao, and X.~Wang, ``Deepreid: Deep filter pairing neural
  network for person re-identification,'' in \emph{CVPR}, 2014.

\bibitem{kalayeh2018human}
M.~M. Kalayeh, E.~Basaran, M.~G{\"o}kmen, M.~E. Kamasak, and M.~Shah, ``Human
  semantic parsing for person re-identification,'' in \emph{Proceedings of the
  IEEE Conference on Computer Vision and Pattern Recognition}, 2018, pp.
  1062--1071.

\bibitem{wang2018resource}
Y.~Wang, L.~Wang, Y.~You, X.~Zou, V.~Chen, S.~Li, G.~Huang, B.~Hariharan, and
  K.~Q. Weinberger, ``Resource aware person re-identification across multiple
  resolutions,'' in \emph{Proceedings of the IEEE Conference on Computer Vision
  and Pattern Recognition}, 2018, pp. 8042--8051.

\bibitem{wang2018mancs}
C.~Wang, Q.~Zhang, C.~Huang, W.~Liu, and X.~Wang, ``Mancs: A multi-task
  attentional network with curriculum sampling for person re-identification,''
  in \emph{Proceedings of the European Conference on Computer Vision (ECCV)},
  2018, pp. 365--381.

\bibitem{xu2018attention}
J.~Xu, R.~Zhao, F.~Zhu, H.~Wang, and W.~Ouyang, ``Attention-aware compositional
  network for person re-identification,'' in \emph{Proceedings of the IEEE
  Conference on Computer Vision and Pattern Recognition}, 2018, pp. 2119--2128.

\bibitem{zheng2019pyramidal}
F.~Zheng, C.~Deng, X.~Sun, X.~Jiang, X.~Guo, Z.~Yu, F.~Huang, and R.~Ji,
  ``Pyramidal person re-identification via multi-loss dynamic training,'' in
  \emph{Proceedings of the IEEE Conference on Computer Vision and Pattern
  Recognition}, 2019, pp. 8514--8522.

\bibitem{zhang2019densely}
Z.~Zhang, C.~Lan, W.~Zeng, and Z.~Chen, ``Densely semantically aligned person
  re-identification,'' in \emph{Proceedings of the IEEE Conference on Computer
  Vision and Pattern Recognition}, 2019, pp. 667--676.

\bibitem{luo2019spectral}
C.~Luo, Y.~Chen, N.~Wang, and Z.~Zhang, ``Spectral feature transformation for
  person re-identification,'' in \emph{Proceedings of the IEEE International
  Conference on Computer Vision}, 2019, pp. 4976--4985.

\bibitem{wang2019spatial}
G.~Wang, J.~Lai, P.~Huang, and X.~Xie, ``Spatial-temporal person
  re-identification,'' in \emph{Proceedings of the AAAI Conference on
  Artificial Intelligence}, vol.~33, 2019, pp. 8933--8940.

\end{thebibliography}

\end{document}